\documentclass{article} 
\usepackage{iclr2023_conference,times}

\usepackage{amsmath,amsfonts,bm}

\def\eqref#1{equation~\ref{#1}}

\def\1{\bm{1}}

\DeclareMathAlphabet{\mathsfit}{\encodingdefault}{\sfdefault}{m}{sl}
\SetMathAlphabet{\mathsfit}{bold}{\encodingdefault}{\sfdefault}{bx}{n}

\newcommand{\E}{\mathbb{E}}

\DeclareMathOperator*{\argmax}{arg\,max}

\usepackage[utf8]{inputenc} 
\usepackage[T1]{fontenc}    
\usepackage[colorlinks=true,citecolor=brown,linkcolor=blue,urlcolor=blue]{hyperref}
\usepackage{url}            
\usepackage{booktabs}       
\usepackage{amsfonts}       
\usepackage{nicefrac}       
\usepackage{microtype}      
\usepackage{xcolor}         
\usepackage{graphicx}
\usepackage{wrapfig}
\usepackage{enumitem}
\usepackage{bbding}
\usepackage[font=small]{caption}
\usepackage{multirow}
\usepackage{subcaption}
\usepackage{listings}

\usepackage{amsmath}
\usepackage{amssymb}
\usepackage{mathtools}
\usepackage{amsthm}
\usepackage{bm}
\usepackage{bbm}

\definecolor{cb_orange}{RGB}{213,94,0}
\definecolor{cb_green}{RGB}{34,136,51}
\definecolor{sky_blue}{RGB}{204, 238, 255}
\definecolor{cb_purple}{RGB}{170, 51, 119}
\definecolor{cb_red}{RGB}{204, 51, 17}
\definecolor{cb_blue}{RGB}{0, 119, 187}

\newcommand{\updates}[1]{{\color{black}{#1}}}

\newcommand{\proposedshort}{MTGv2}  
\newcommand{\proposedvariant}{MTGv1}  
\newcommand{\proposedrep}{morphology-task graph}
\newcommand{\proposedREP}{Morphology-Task Graph}
\newcommand{\proposedRep}{Morphology-task graph}
\newcommand{\proposedmainshort}{MTG}

\newcommand{\proposedbench}{MxT-Bench}

\title{\updates{A System} for Morphology-Task Generalization \updates{via Unified Representation and Behavior Distillation}}

\author{
Hiroki Furuta$^{1,2}$\thanks{Work done as Student Researcher at Google.} \quad
Yusuke Iwasawa$^{1}$ \quad
Yutaka Matsuo$^{1}$ \quad
Shixiang Shane Gu$^{2,1}$ \\
$^{1}$The University of Tokyo \quad
$^{2}$Google Research, Brain Team\\
\texttt{furuta@weblab.t.u-tokyo.ac.jp} \\
}

\iclrfinalcopy 

\begin{document}

\maketitle

\begin{abstract}

The rise of generalist large-scale models in natural language and vision has made us expect that a massive data-driven approach could achieve broader generalization in other domains such as continuous control. In this work, we explore a method for learning a single policy that manipulates various forms of agents to solve various tasks by distilling a large amount of proficient behavioral data. In order to align input-output (IO) interface among multiple tasks and diverse agent morphologies while preserving essential 3D geometric relations, we introduce \textit{\proposedrep{}}, which treats observations, actions and goals/task in a unified graph representation.
We also develop \proposedbench{} for fast large-scale behavior generation, which supports procedural generation of diverse morphology-task combinations with a minimal blueprint and hardware-accelerated simulator.
Through efficient representation and architecture selection on \proposedbench{}, we find out that a \proposedrep{} representation coupled with Transformer architecture improves the multi-task performances compared to other baselines including recent discrete tokenization, and provides better prior knowledge for zero-shot transfer or sample efficiency in downstream multi-task imitation learning.
Our work suggests large diverse offline datasets, unified IO representation, and policy representation and architecture selection through supervised learning form a promising approach for studying and advancing morphology-task generalization\footnote{\url{https://sites.google.com/view/control-graph}}.
\end{abstract}

\section{Introduction}

The impressive success of large language models~\citep{devlin2019bert,radford2019language,bommasani2021opportunities,brown2020language,Chowdhery2022palm} has encouraged the other domains, such as computer vision~\citep{radford2021clip,gu2021vild,Alayrac2022flamingo,jaegle2021perceiver} or robotics~\citep{Ahn2022saycan,huang2022inner}, to leverage the large-scale pre-trained model trained with massive data with unified input-output interface.
These large-scale pre-trained models are innately multi-task learners: they surprisingly work well not only in the fine-tuning or few-shot transfer but also in the zero-shot transfer settings~\citep{2020t5,chen2022pali}.
Learning a ``generalist'' model seems to be an essential goal in the recent machine learning paradigm with the same key ingredients: curate \textbf{massive diverse dataset}, define \textbf{unified IO representation}, and perform \textbf{efficient representation and architecture selection}, altogether for best generalization.

In reinforcement learning (RL) for continuous control, various aspects are important for generalization.
First, we care about ``task'' generalization. For instance, in robotic manipulation, we care the policy to generalize for different objects and target goal positions~\citep{kalashnikov2018scalable,andrychowicz2017hindsight,yu2019meta,lynch2019learning}. Recent advances in vision and language models also enable task generalization through compositional natural language instructions~\citep{jiang2019language,shridhar2022cliport,Ahn2022saycan,cui2022can}.
However, to scale the data, equally important is ``morphology'' generalization, where a single policy can control agents of different embodiment~\citep{wang2018nervenet,noguchi2021tool} and can thereby ingest experiences from as many robots in different simulators~\citep{brax2021github,todorov2012mujoco,coumans2016pybullet} as possible.
Most prior works~\citep{mendonca2021lexa,gupta2022metamorph} only address either the task or morphology axis separately, and achieving broad generalization over task and morphology jointly remains a long-standing problem\footnote{\updates{Here, ``task'' means what each agent should solve. See Section~\ref{sec:preliminaries} for the detailed definition. When considering the models, as slightly overloaded, it may imply morphological diversity as well.}}.


\begin{figure*}[t]
    \centering
    \includegraphics[width=0.85\linewidth]{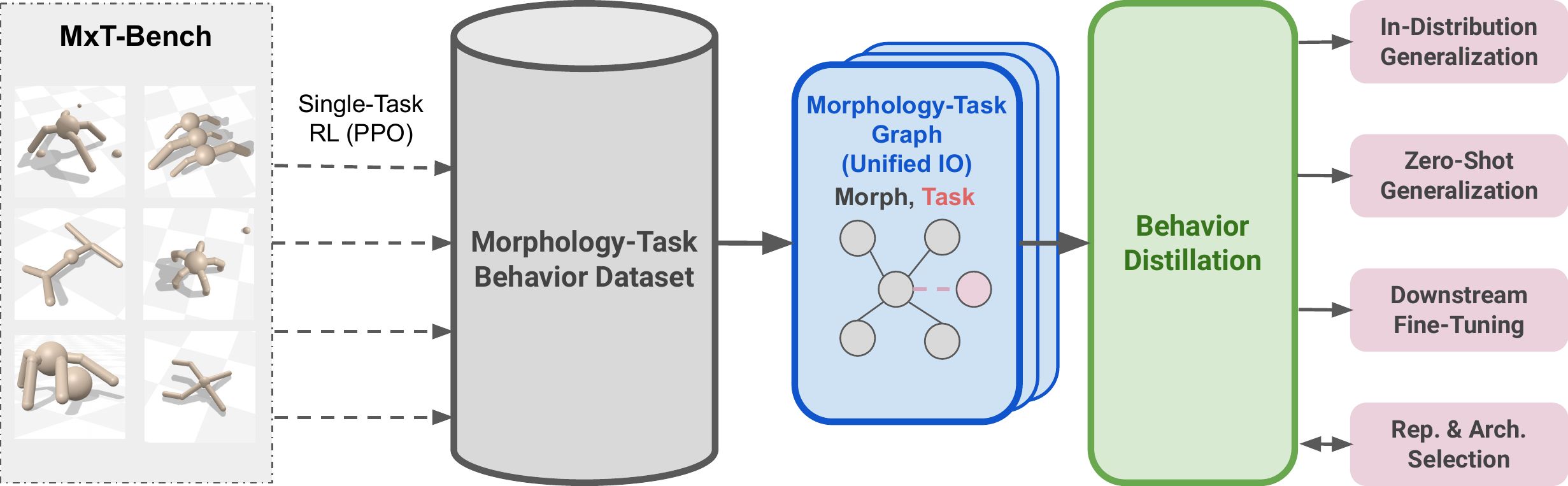}
    \vspace{-5pt}
    \caption{
    Behavior distillation pipeline. We first train a single-task policy for each environment on \proposedbench{}, and then collect proficient morphology-task behavior dataset\updates{~(Section~\ref{sec:embodiment_bench})}.
    To enable a single policy to learn multiple tasks and morphologies simultaneously, we convert stored transitions to the \proposedrep representation to align with unified IO interface\updates{~(Section~\ref{sec:graph_rep_bench})} for multi-task distillation\updates{~(Section~\ref{sec:behavior_distillation})}. After behavior distillation, the learned policy can be utilized for in-distribution or zero-shot generalization\updates{~(Section~\ref{sec:distillation_results})}, downstream fine-tuning\updates{~(Section~\ref{sec:zero_shot_finetuning_results})}, and representation and architecture selection\updates{~(Section~\ref{sec:node_feature_selection})}.
    }
    \label{fig:behavior_distillation}
    \vskip -0.1in
\end{figure*}

This paper first proposes \textit{\proposedbench{}}\footnote{Pronounced as ``mixed''-bench. It stands for ``Morphology $\times$ Task''.}, the first multi-morphology and multi-task benchmarking environments, as a step toward building the \textbf{massive diverse dataset} for continuous control.
\proposedbench{} provides various combinations of different morphologies (ant, centipede, claw, worm, and unimal~\citep{gupta2022metamorph}) and different tasks (reach, touch, and twisters). 
\proposedbench{} is easily scalable to additional morphologies and tasks, and is built on top of Brax~\citep{brax2021github} for fast behavior generation.

Next, we define \textbf{unified IO representation} for an architecture to ingest all the multi-morphology multi-task data.
Inspired by \textit{scene graph}~\citep{johnson2015scene} in computer vision that represents the 3D relational information of a scene, and by \textit{morphology graph}~\citep{wang2018nervenet,chen2018hardware,huang2020policy,gupta2022metamorph} that expresses an agent's geometry and actions, 
we introduce the notion of \textit{\proposedrep{} (\proposedmainshort{})} as a unified interface to encode
observations, actions, and goals (i.e. tasks) as nodes in the shared graph representation.
Goals are represented as sub-nodes, and different tasks correspond to different choices: touching is controlling a torso node, while reaching is controlling an end-effector node (\autoref{fig:graph_representation}).
In contrast to discretizing and tokenizing every dimension as proposed in recent work~\citep{janner2021sequence,reed2022gato}, this unified IO limits data representation it can ingest, but strongly preserves 3D geometric relationships that are crucial for any physics control problem~\citep{wang2018nervenet,ghasemipour22blocks}, and we empirically show it outperforms naive tokenization in our control-focused dataset.

Lastly, while conventional multi-task or meta RL studies generalization through on-policy joint training~\citep{yu2019meta,cobbe2020leveraging}, we perform \textbf{efficient representation and architecture selection}, over 11 combinations of unified IO representation and network architectures, and 8 local node observations,
for optimal generalization through \textit{behavior distillation} (\autoref{fig:behavior_distillation}), where RL is essentially treated as a (single-task, low-dimensional) behavior generator~\citep{gu2021braxlines} and multi-task supervised learning (or offline RL~\citep{fujimoto2019off}) is used for imitating all the behaviors~\citep{singh2021parrot,chen2021a,reed2022gato}.
Through offline distillation, we controllably and tractably evaluate two variants of \proposedmainshort{} representation, along with multiple network architectures (MLP, GNN~\citep{kipf2017semi}, Transformers~\citep{vaswani2017attention}), and show that
\proposedshort{} variant with Transformer improves the multi-task goal-reaching performances compared to other possible choices by 23\% and provides better prior knowledge for zero-shot generalization (by 14$\sim$18\%) and fine-tuning for downstream multi-task imitation learning (by 50 $\sim$ 55 \%).

As the fields of vision and language move toward broad generalization~\citep{chollet2019measure,bommasani2021opportunities}, we hope our work could encourage RL and continuous control communities to continue growing diverse behavior datasets, designing different IO representations, and iterating more representation and architecture selection, and eventually optimize a single policy that can be deployed on any morphology for any task. 
In summary, our key contributions are:
\begin{itemize}[leftmargin=1.0cm,topsep=0pt,itemsep=0.5pt]
     \item We develop \textit{\proposedbench}\footnote{
     \url{https://github.com/frt03/mxt_bench}
     } as a test bed for \textit{morphology-task generalization} with fast expert behavior generator.
     \proposedbench{} supports the scalable procedural generation of both agents and tasks with minimal blueprints.
    \item We introduce \textit{\proposedrep}, a universal IO for control which treats the agent's observations, actions and goals/tasks in a unified graph representation, while preserving the task structure.
    \item We study generalization through offline supervised behavior distillation, where we can efficiently try out various design choices; over 11 combinations of unified IO representation and network architectures, and 8 local node observations. As a result, we find that Transformer with \proposedshort{} achieves the best multi-task performances among other possible designs (MLP, GNN and Transformer with \proposedvariant{}, Tokenized-\proposedshort{}, etc.) in both in-distribution and downstream tasks, such as zero-shot transfer and fine-tuning for multi-task imitation learning.
\end{itemize}

\begin{wrapfigure}{R}[0pt]{0.4\linewidth}
    \centering
    \includegraphics[width=\linewidth]{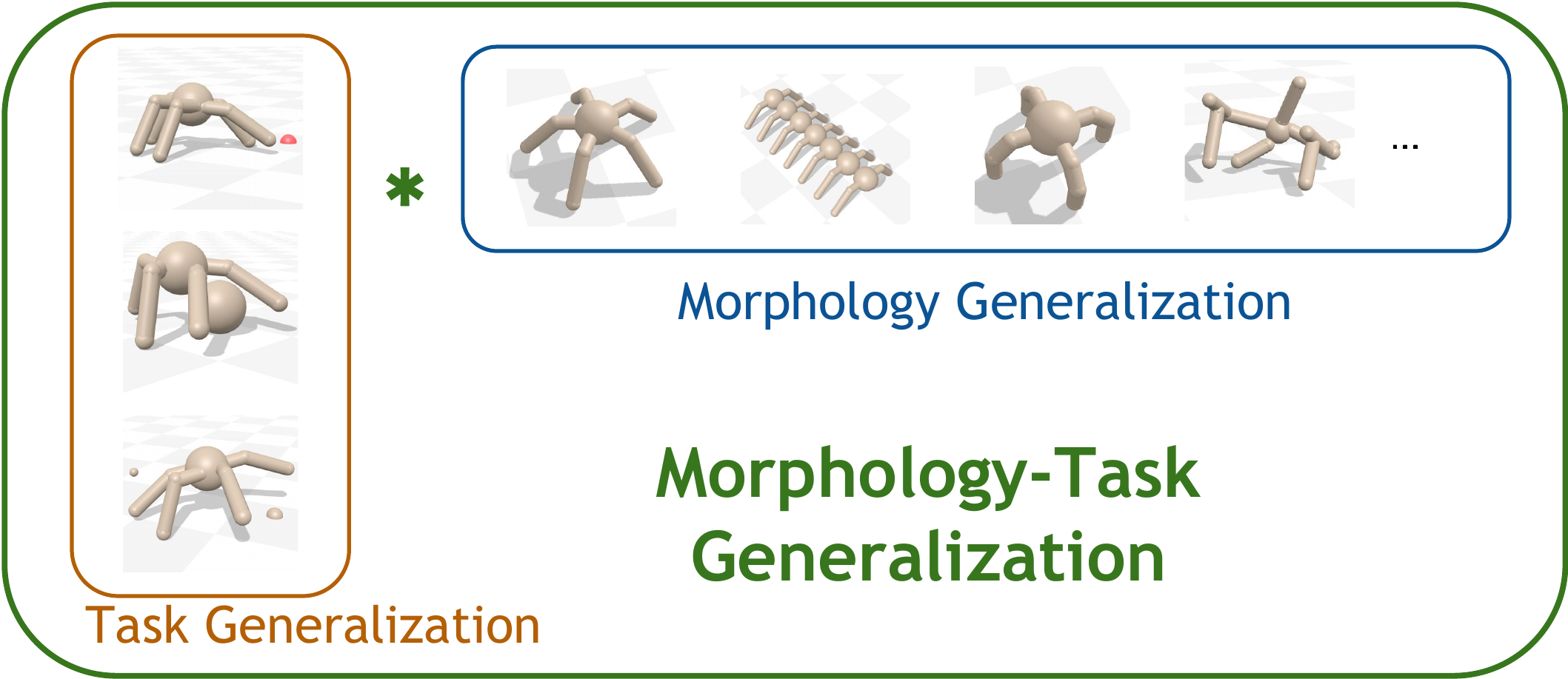}
    \vspace{-10pt}
    \caption{We tackle \textit{morphology-task generalization}, which requires achieving both task and morphology generalization simultaneously.
    \updates{See \autoref{sec:extended_related_work} for the details.}
    }
    \vskip -0.1in
    \label{fig:embodiment_challenge}
\end{wrapfigure}

\section{Related Work}
\label{sec:related}

\textbf{Morphology Generalization}~
While, in RL for continuous control, the policy typically learns to control only a single morphology~\citep{tassa2018deepmind,todorov2012mujoco}, 
several works succeed in generalizing the control problem for morphologically different agents to solve a locomotion task by using morphology-aware Graph Neural Network (GNN) policies~\citep{wang2018nervenet,huang2020policy,blake2021snowflake}. 
In addition, several work~\citep{kurin2020my,gupta2022metamorph, hong2022structureaware,br2022anymorph} have investigated the use of Transformer~\citep{vaswani2017attention}.
Other work jointly optimize the morphology-agnostic policy and morphology itself~\citep{pathak19assemblies,gupta2021embodied,yuan2022transformact,hejna2021tame}, or transfer a controller over different morphologies~\citep{devin2017modular,chen2018hardware,hejna2020hierarchically,liu2022revolver}.

While substantial efforts have been investigated to realize morphology generalization, those works mainly focus on only a single task (e.g. running), and less attention is paid to multi-task settings, where the agents attempt to control different states of their bodies to desired goals.
We believe that goal-directed control is a key problem for an embodied single controller.
Concurrently, \citet{feng2022genloco} propose an RL-based single controller that is applied to different quadruped robots and target poses in the sim-to-real setting.
In contrast, our work introduces the notion of \proposedrep{} as a unified IO that represents observations, actions, and goals in a shared graph, and can handle more diverse morphologies to solve multiple tasks.

\textbf{Task Generalization}~
In the previous research, task generalization has been explored in multi-task or meta RL literature~\citep{wang2016metarl,duan2017rl,cabi2017intent,teh2017distral,colas2019curious,li2020generalized,yang2020multitask,kurin2022indefence}.
Each task might be defined by the difference in goals, reward functions, and dynamics~\citep{ghasemipour2019smile,kalashnikov2021mtopt,eysenbach2020rewriting}, under shared state and action spaces.
Some works leverage graph representation to embed the compositionality of manipulation tasks~\citep{li19relationalrl,zhou2022policy,li2021solving,kumar2022inverse,ghasemipour22blocks}, while others use natural language representation to specify diverse tasks~\citep{jiang2019language,shridhar2022cliport,Ahn2022saycan,huang2022language,cui2022can}.
Despite the notable success of acquiring various task generalization, multi-task RL often deals with only a single morphology.
We aim to extend the general behavior policy into the ``cartesian product'' of tasks and morphologies~(as shown in \autoref{fig:embodiment_challenge}) to realize a more scalable and capable controller. 

\textbf{Transformer for RL}~
Recently, \citet{chen2021decision} and \citet{janner2021sequence} consider offline RL as supervised sequential modeling problem and following works achieve impressive success~\citep{reed2022gato,lee2022mgdt,furuta2021generalized,xu2022prompt,Shafiullah2022behavior,zheng2022online,paster2022youcant}.
In contrast, our work leverages Transformer to handle topological and geometric information of the scene, rather than a sequential nature of the agent trajectory. 

\textbf{Behavior Distillation}~
Due to the massive try-and-error and large variance, training a policy from scratch in online RL is an inefficient process, especially in multi-task setting.
It is more efficient to use RL for generating single-task behaviors (often from low dimensions)~\citep{gu2021braxlines} and then use supervised learning to imitate all behaviors with a large single policy~\citep{levine2016end,rusu2016policy,parisotto2016actor,singh2021parrot,ajay2021opal,chen2021a}.
Several works tackle the large-scale behavior distillation with Transformer~\citep{reed2022gato,lee2022mgdt}, or with representation that treats observations and actions in the same vision-language space~\citep{zeng2020transporter,shridhar2022perceiveractor}.
Our work utilizes similar pipeline, but focuses on finding the good representation and architecture to generalize across morphology and tasks simultaneously with proposed \proposedrep{}.
\updates{See \autoref{sec:extended_related_work} for the connection to policy distillation.}

\section{Preliminaries}
\label{sec:preliminaries}
In RL, consider a Markov Decision Process with following tuple ($\mathcal{S}$, $\mathcal{A}$, $p$, $p_1$, $r$, $\gamma$), which consists of state space $\mathcal{S}$, action space $\mathcal{A}$, state transition probability function $p: \mathcal{S} \times \mathcal{A} \times \mathcal{S} \rightarrow [0, \infty)$, initial state distribution $p_1 : \mathcal{S} \rightarrow [0, \infty)$, reward function $r: \mathcal{S} \times \mathcal{A} \rightarrow \mathbb{R}$, and discount factor $\gamma \in [0 ,1)$.
\updates{The agent follows a Markovian policy $\pi: \mathcal{S} \times \mathcal{A} \rightarrow [0, \infty)$, which is often parameterized in Deep RL, and} seeks optimal policy $\pi^{*}$ that maximizes the discounted cumulative rewards:
\begin{equation}
    \textstyle
    \pi^{*} = \underset{\pi}{\argmax}~ \frac{1}{1-\gamma} \E_{s \sim \rho^\pi(s), a\sim \pi(\cdot|s)}\left[ r(s, a)\right], \label{eq:rl}
\end{equation}
where $p_t^\pi(s_{t})=\iint_{s_{0:t}, a_{0:t-1}} \prod_t p(s_t|s_{t-1},a_{t-1})\pi(a_t|s_t)$ and $\rho^\pi(s) = (1-\gamma) \sum_{t} \gamma^{t} p_{t}^\pi(s_{t}=s)$ represent time-aligned and time-aggregated state marginal distributions following policy $\pi$.

\begin{figure*}[t]
    \centering
    \includegraphics[width=0.8\linewidth]{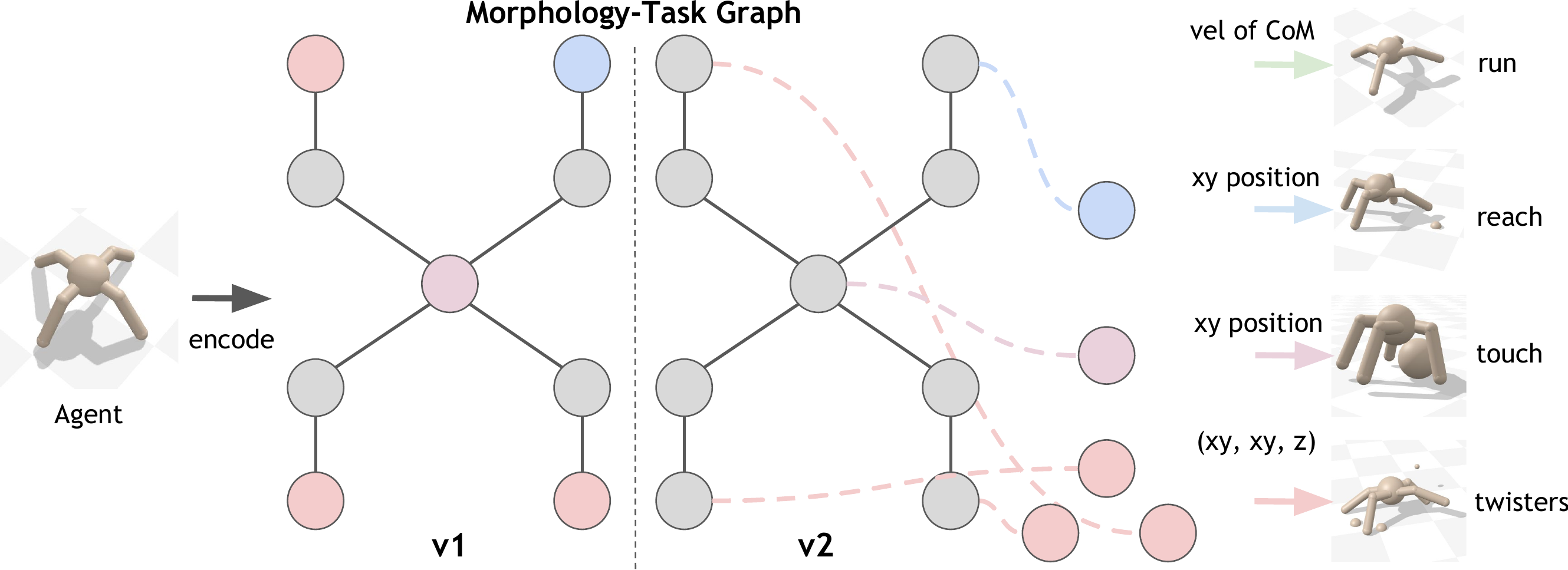}
    \vspace{-5pt}
    \caption{
    We propose the notion of \textit{\proposedrep{}}, which expresses the agent's observations, actions, and goals/tasks in a unified graph representation, while preserving the geometric structure of the task.
    We develop two practical implementations; \textbf{\proposedrep{} v1} (left) accepts the morphological graph, encoded from the agent's geometric information, as an input-output interface, and merges positional goal information as a part of corresponding node features.
    \textbf{\proposedrep{} v2} (right) treats given goals as extra disjoint nodes of the morphological graph.
    While most prior morphology-agnostic RL have focused on locomotion (\textcolor{cb_green}{run} task), i.e. a \textcolor{cb_green}{single static goal node} controlling (maximizing) the \textcolor{cb_green}{velocity} of center of mass, \proposedrep{} could naturally extend morphology-agnostic control to other goal-oriented tasks: \textcolor{cb_blue}{single static goal node} for \textcolor{cb_blue}{reach},  \textcolor{cb_red}{multiple static goal nodes} for \textcolor{cb_red}{twisters}, and \textcolor{cb_purple}{single dynamic goal node} tracking a movable ball for an object interaction task; \textcolor{cb_purple}{touch}.
}
    \label{fig:graph_representation}
    \vskip -0.1in
\end{figure*}

\textbf{Graph Representation for Morphology-Agnostic Control}~
Following prior continuous control literature~\citep{todorov2012mujoco}, we assume the agents have bodies modeled as simplified skeletons of animals.
An agent's morphology is characterized by the parameter for rigid body module (torso, limbs), such as radius, length, mass, and inertia, and by the connection among those modules (joints).
In order to handle such geometric and topological information, an agent's morphology can be expressed as an acyclic tree graph representation $\mathbf{G} := (\mathbf{V}, \mathbf{E})$, where $\mathbf{V}$ is a set of nodes $v^{i} \in \mathbf{V}$ and $\mathbf{E}$ is a set of edges $e^{ij} \in \mathbf{E}$ between $v^{i}$ and $v^{j}$.
The node $v^{i}$ corresponds to $i$-th module of the agent, and the edge $e^{ij}$ corresponds to the hinge joint between the nodes $v^{i}$ and $v^{j}$.
Each joint may have 1-3 actuators corresponding to a degree of freedom. If a joint has several actuators, the graph $\mathcal{G}$ is considered a multipath one.
This graph-based \updates{factored} formulation can describe the various agents' morphologies in a tractable manner~\citep{wang2018nervenet,huang2020policy,loynd2020working}.

We assume that node $v^{i}$ observes local sensory input $s_{t}^{i}$ at time step $t$, which includes the information of limb $i$ such as position, velocity, orientation, joint angle, or morphological parameters.
To process these node features and graph structure, a morphology-agnostic policy can be modeled as node-based GNN~\citep{kipf2017semi,battaglia2018gnn,cappart2021combinational}, which takes a set of local observations $\{s_{t}^{i}\}_{i=1}^{|\mathbf{V}|}$
as an input and emit the actions for actuators of each joint $\{a_{t}^{e}\}_{e=1}^{|\mathbf{E}|}$.
The objective of morphology-agnostic RL is the average of \autoref{eq:rl} among given morphologies.

\textbf{Goal-conditional RL}~
In goal-conditional RL~\citep{Kaelbling93learning,schaul2015uvf}, the agent aims to find an optimal policy $\pi^{*}(a|s, s_g)$ conditioned on goal $s_g \in \mathcal{G}$, where $\mathcal{G}$ stands for goal space that is sub-dimension of state space $\mathcal{S}$ (e.g. XYZ coordinates, velocity, or quaternion). The desired goal $s_g$ is sampled from the given goal distribution $p_{\psi}: \mathcal{G} \rightarrow [0, \infty)$, where $\psi$ stands for task in the task space $\Psi$ (e.g. reaching the agent's leg or touching the torso to the ball).
The reward function can include a goal-reaching term, that is often modeled as $r_{\psi}(s_t, s_g) = -d_{\psi}(s_t, s_g)$, where $d_{\psi}(\cdot, \cdot)$ is a task-dependent distance function, such as Euclidean distance, between the sub-dimension of interest in current state $s_t$ and given goal $s_g$. 
Some task $\psi$ give multiple goals to the agents. In that case, we overload $s_g$ to represent a set of goals; $\{s_g^{i}\}_{i=1}^{N_{\psi}}$, where $N_{\psi}$ is the number of goals that should be satisfied in the task $\psi$.

\textbf{Morphology-Task Generalization}~
This paper aims to achieve morphology-task generalization, where the learned policy should generalize over tasks and morphologies simultaneously. 
The optimal policy should generalize over morphology space $\mathcal{M}$, task $\Psi$, and minimize the distance to any given goal $ s_g \in \mathcal{G}$.
Mathematically, this objective can be formulated as follows:
\begin{equation}
    \textstyle
    \pi^{*} = \underset{\pi}{\argmax}~ \frac{1}{1-\gamma} \E_{m,\psi \sim \mathcal{M}, \Psi} \left[ \E_{s_g \sim p_{\psi}(s_g)} \left[ \E_{\bm{s}^m, \bm{a}^m \sim \rho^\pi(\bm{s}^m), \pi(\cdot|\bm{s}^m, s_g)}\left[ -d_{\psi}(\bm{s}^m, s_g) \right] \right] \right], \label{eq:egrl}
\end{equation}
where the graph representation of morphology $m \in \mathcal{M}$ is denoted as $\mathbf{G}_{m} = (\mathbf{V}_{m}, \mathbf{E}_{m})$, and $\bm{s}^m := \{s_{t}^{i}\}_{i=1}^{|\mathbf{V}_{m}|}$ and $\bm{a}^m := \{a_{t}^{e}\}_{e=1}^{|\mathbf{E}_{m}|}$ stand for a set of local observations and actions of morphology $m$.
While we can use multi-task online RL to maximize \autoref{eq:egrl} in principle, it is often sample inefficient due to the complexity of task, which requires a policy that can handle the diversity of the scene among morphology $\mathcal{M}$, task $\Psi$, and goal space $\mathcal{G}$ simultaneously.

\section{Method}
\subsection{\proposedbench{} as a Test Bed for Morphology-Task Generalization}
\label{sec:embodiment_bench}

\begin{figure*}[t]
\begin{minipage}[c]{0.5\textwidth}
    \centering
    \includegraphics[width=\linewidth]{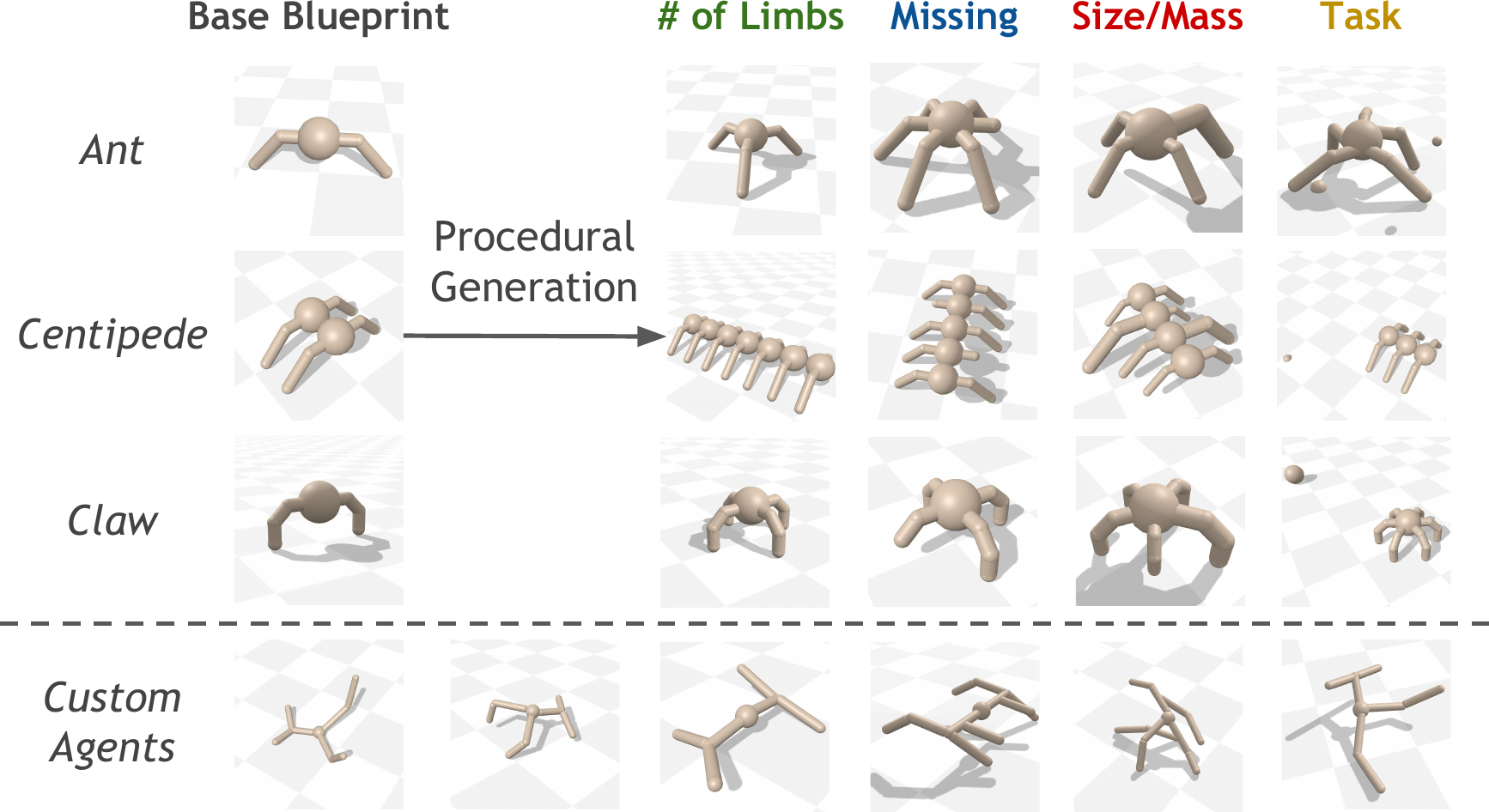}
\end{minipage}
\begin{minipage}[t]{0.475\textwidth}
\begin{center}
    \scalebox{0.55}{
    \begin{tabular}{l|ccc}
        \toprule
        \textbf{Benchmark} & \textbf{Multi-Task} & \textbf{Multi-Morphology} & \textbf{Scalability} \\
        \midrule
        MuJoCo \& DM Control & \textcolor{red}{\XSolidBrush} & \textcolor{red}{\XSolidBrush} & \textcolor{red}{\XSolidBrush} \\
        Meta-World~\citep{yu2019meta} & \textcolor{green}{\CheckmarkBold} & \textcolor{red}{\XSolidBrush} & \textcolor{red}{\XSolidBrush} \\
        \citet{huang2020policy} & \textcolor{red}{\XSolidBrush} & \textcolor{green}{\CheckmarkBold} &  \textcolor{red}{\XSolidBrush}\\
        \citet{gupta2022metamorph} & \textcolor{red}{\XSolidBrush} & \textcolor{green}{\CheckmarkBold} & \textcolor{green}{\CheckmarkBold} \\
        \textbf{\proposedbench (Ours)} & \textcolor{green}{\CheckmarkBold} & \textcolor{green}{\CheckmarkBold} & \textcolor{green}{\CheckmarkBold} \\
        \bottomrule
    \end{tabular}
    }
\end{center}
\end{minipage}
\vskip -0.1in
\caption{
The overview of \proposedbench{}, which can procedurally generate both various morphologies and tasks with minimal blueprints.
\proposedbench{} can not only construct the agents with different number of limbs, but also randomize missing limbs and size/mass of bodies.
We could design the tasks with parameterized goal distributions.
It also supports to import custom complex agents such as unimals~\citep{gupta2022metamorph}.
Compared to relevant RL benchmarks in terms of (1) multi-task (task coverage), (2) multi-morphology (morphology coverage), and (3) scalability, MuJoCo~\citep{todorov2012mujoco} and DM Control~\citep{tassa2018deepmind} only have a single morphology for a single task. While other existing works~\citep{yu2019meta,huang2020policy} partially cover task-/morphology-axis with some sort of scalability, they do not satisfy all criteria.
}
\label{fig:embodiment_bench}
\vskip -0.1in
\end{figure*}

To overcome these shortcomings in the existing RL environments, we develop \textit{\proposedbench}, which has a wide coverage over both tasks and morphologies to test morphology-task generalization, with the functionalities for procedural generation from minimal blueprints~(\autoref{fig:embodiment_bench}).
\proposedbench{} is built on top of Brax~\citep{brax2021github} and Composer~\citep{gu2021braxlines}, for faster iteration of behavior distillation with hardware-accelerated environments.
Beyond supporting multi-morphology and multi-task settings, the scalability of \proposedbench{} helps to test the broader range of morphology-task generalization since we can easily generate out-of-distribution tasks and morphologies, compared to manually-designed morphology or task specifications.

In the morphology axis, we prepare 4 types of blueprints (ant, claw, centipede, and worm) as base morphologies, since they are good at the movement on the XY-plane.
Through \proposedbench, we can easily spawn agents that have different numbers of bodies, legs, or different sizes, lengths, and weights.
Moreover, we can also import the existing complex morphology used in previous work. For instance, we include 60+ morphologies that are suitable for goal-reaching, adapted from \citet{gupta2022metamorph} designed in MuJoCo.
In the task axis, we design reach, touch, and twisters~\footnote{\url{https://en.wikipedia.org/wiki/Twister_(game)}} as basic tasks, which could evaluate different aspects of the agents; the simplest one is the reach task, where the agents aim to put their leg on the XY goal position. 
In the touch task, agents aim to create and maintain contact between a specified torso and a movable ball.
The touch task requires reaching behavior, while maintaining a conservative momentum to avoid kicking the ball away from the agent. Twisters tasks are the multi-goal problems; for instance, the agents should satisfy XY-position for one leg, and Z height for another leg.
We pre-define 4 variants of twisters with max 3 goals (see Appendix~\ref{sec:mxt_bench_task} for the details).
Furthermore, we could easily specify both initial and goal position distribution with parameterized distribution. 
In total, we prepare 180+ environments combining the morphology and task axis for the experiments in the later section.
See \autoref{sec:embodiment_bench_details} for further details.

\subsection{Behavior Distillation}
\label{sec:behavior_distillation}
Toward broader generalization over morphologies and tasks, a single policy should handle the diversity of the scene among morphology $\mathcal{M}$, task $\Psi$, and goal space $\mathcal{G}$ simultaneously.
Multi-task online RL from scratch, however, is difficult to tune, slow to iterate, and hard to reproduce.
Instead, we employ behavior cloning on RL-generated expert behaviors
to study morphology-task generalization.
To obtain rich goal-reaching behaviors, we train a single-morphology single-task policy using PPO~\citep{Schulman2017PPO} with a simple MLP policy, which is significantly more efficient than multi-morphology training done in prior work~\citep{gupta2022metamorph,kurin2020my,huang2020policy}.
Since \proposedbench{} is built on top of Brax~\citep{brax2021github}, a hardware-accelerated simulator, training PPO policies can be completed in about 5$\sim$30 minutes per environment (on NVIDIA RTX A6000). 
We collect many behaviors per morphology-task combination from the expert policy rollout.
We then train a single policy $\pi_{\theta}$ with a supervised learning objective:
\begin{equation}
    \textstyle
    \mathcal{L}_{\pi} = - \E_{m, \psi \sim \mathcal{M}, \Psi} \left[\E_{\bm{s}^m, \bm{a}^m, s_g \sim \mathcal{D}_{m, \psi}} \left[\log \pi_{\theta}(\bm{a}^m|\{\bm{s}^m, s_g\})\right]\right], \nonumber
\end{equation}
where $\mathcal{D}_{m, \psi}$ is an expert dataset of morphology $m$ and task $\psi$.
Importantly, offline behavior distillation protocol runs (parallelizable) single-task online RL only once, and allows us to reuse the same fixed data to try out various design choices, such as model architectures or local features of \proposedrep{}, which is often intractable in multi-task online RL.

\subsection{\proposedREP}
\label{sec:graph_rep_bench}
To learn a single policy that could solve various morphology-task problems, it is essential to unify the input-output interface among those.
Inspired by the concept of scene graph~\citep{johnson2015scene} and morphological graph~\citep{wang2018nervenet}, we introduce the notion of \proposedrep{} (\proposedmainshort) representation that incorporates goal information while preserving the geometric structure of the task.
\proposedRep{} could express the agent's observations, actions, and goals/tasks in a unified graph space.
Although most prior morphology-agnostic RL has focused on locomotion (running) with the reward calculated from the velocity of the center of mass, \proposedrep{} could naturally extend morphology-agnostic RL to multi-task goal-oriented settings: including static single positional goals (reaching), multiple-goal problems (twister-game) and object interaction tasks (ball-touching).
In practice, we develop two different ways to inform the goals and tasks~(\autoref{fig:graph_representation}); \proposedrep{} v1 (\proposedvariant{}) accepts the morphological graph, encoded from the agent's geometric information, as an input-output interface, and merges positional goal information as a part of corresponding node features.
For instance, in touch task, \proposedvariant{} includes XY position of the movable ball as an extra node feature of the body node.
Moreover, \proposedrep{} v2 (\proposedshort{}) considers given goals as additional disjoint nodes of morphological graph representation.
These \proposedrep{} strategies enable the policy to handle a lot of combinations of tasks and morphologies simultaneously.

\textbf{Transformer for \proposedmainshort{}}~
While \proposedrep{} could represent the agent and goals in a unified manner, because the task structure may change over time, the policy should unravel the implicit relationship between the agent's modules and goals dynamically. We mainly employ Transformer as a policy architecture because it can process morphological graph as a fully-connected graph and achieve notable performance by leveraging the hidden relations between the nodes beyond manually-encoded geometric structure~\citep{kurin2020my}.
\updates{The combination of \proposedrep{} and Transformer is expected to model the local relations between goals and target nodes explicitly and to embed the global relations among the modules.} 
See~\autoref{sec:implementation} for further details. 

\begin{table*}[t]
\begin{center}
\begin{small}
\scalebox{0.675}{
\begin{tabular}{l|cccccc}
\toprule
 & \textbf{Random} & \textbf{MLP} & \textbf{GNN (\proposedvariant{})} & \textbf{Transformer (\proposedvariant{})} & \textbf{Transformer (\proposedshort{})} & \textbf{Token-\proposedshort{}}\\
\midrule
\textbf{In-Distribution} & 1.2019 $\pm$ 0.41 & 0.5150 $\pm$ 0.01 & 0.4776 $\pm$ 0.01 & 0.4069 $\pm$ 0.02 & \textbf{0.3128 $\pm$ 0.02} & 0.3402 $\pm$ 0.01\\
\textbf{In-Distribution (unimal)} & 0.9090 $\pm$ 0.03 & 0.6703 $\pm$ 0.01 & -- & 0.4839 $\pm$ 0.02 & \textbf{0.4178 $\pm$ 0.01} & -- \\
\midrule
\textbf{Compositional (Morphology)} & 1.1419 $\pm$ 0.41 & 0.7216 $\pm$ 0.01 & -- & 0.4940 $\pm$ 0.01 & \textbf{0.4066 $\pm$ 0.01} & -- \\
\textbf{Compositional (Task)} & 0.8932 $\pm$ 0.01 & 0.6849 $\pm$ 0.01  & -- & 0.5395 $\pm$ 0.04 & \textbf{0.4461 $\pm$ 0.05} & -- \\
\textbf{Out-of-Distribution} & 0.8979 $\pm$ 0.01 & 0.7821 $\pm$ 0.02 & -- & 0.6144 $\pm$ 0.04 & \textbf{0.5266 $\pm$ 0.04} & -- \\
\bottomrule
\end{tabular}
}
\end{small}
\end{center}
\vskip -0.1in
\caption{The average normalized final distance in various types of morphology-task generalization on \proposedbench{}.
We test (1) in-distribution, (2) compositional morphology/task, and (3) out-of-distribution generalization.
(2) and (3) evaluate zero-shot transfer.
We compare MLP, GNN and Transformer with \proposedvariant{}, Transformer with \proposedshort{}, and tokenized \proposedshort{}~\citep{reed2022gato}.
\proposedshort{} improves multi-task performance to other choices by 23 \% in the in-distribution evaluation, and achieves better zero-shot transfer in the compositional and out-of-distribution settings by 14 $\sim$ 18 \%.
}
\label{tab:distillation}
\vskip -0.1in
\end{table*}

\section{Experiments}
\label{sec:experiments}
We first evaluate the multi-task performance of \proposedrep{} representation (\proposedvariant{}, \proposedshort{}) in terms of in-distribution (known morphology and task with different initialization), compositional (known task with unseen morphology or known morphology with unseen task), and out-of-distribution generalization (either morphology or task is unseen) on \proposedbench{}~(Section~\ref{sec:distillation_results}).
Then, we investigate whether \proposedrep{} could contribute to obtaining better control prior for multi-task fine-tuning~(Section~\ref{sec:zero_shot_finetuning_results}).
\updates{In addition, we conduct the offline node feature selection to identify what is the most suitable node feature set for morphology-task generalization~(Section~\ref{sec:node_feature_selection}).}
The results are averaged among 4 random seeds.
See \autoref{sec:implementation} for the hyperparameters.
We also investigate the other axis of representations or architectures~(\autoref{sec:gato_representtion}, \ref{sec:pe_results}, \ref{sec:temporal_history})
and test the effect of dataset size, the number of morphology-task combinations, and model size (\autoref{sec:data_entry}).
\updates{Lastly, we examine why \proposedrep{} works well by visualizing attention weights~(\autoref{sec:attention_details}).}

\textbf{Evaluation Metric}~
Goal-reaching tasks are evaluated by the distance to the goals at the end of episode~\citep{pong2018temporal,pong2020skew,ghosh2021learning,choi2021variational,eysenbach2021clearning}.
However, this can be problematic in our settings, because the initial distance or the degree of goal-reaching behaviors might be different among various morphologies and tasks.
We measure the performance of the policy $\pi$ by using a normalized final distance metric $\bar{d}(\mathcal{M}, \Psi; \pi)$ over morphology $\mathcal{M}$ and task space $\Psi$ with pre-defined max/min value of each morphology $m$ and task $\psi$,
\begin{equation}
    \textstyle
    \bar{d}(\mathcal{M}, \Psi; \pi) := \frac{1}{|\mathcal{M}||\Psi|} \sum_{m}^{\mathcal{M}} \sum_{\psi}^{\Psi} \E_{s_g \sim p_{\psi}} \left[ \sum_{i=1}^{N_{\psi}} \frac{d_{\psi}(\bm{s}^m_T, s_g^{i}) - d_{\min}^{i,m,\psi}}{d_{\max}^{i,m,\psi} - d_{\min}^{i,m,\psi}} \right] ,
    \label{eq:normalized_dist}
\end{equation}
where $\bm{s}^m_T$ is the last state of the episode, $d_{\max}^{i,m,\psi}$ is a maximum, and $d_{\min}^{i,m,\psi}$ is a minimum distance of $i$-th goal $s_g^{i}$ with morphology $m$ and task $\psi$.
We use a distance threshold to train the expert PPO policy as $d_{\min}^{i,m,\psi}$, and the average distance from an initial position of the scene as $d_{\max}^{i,m,\psi}$.
\autoref{eq:normalized_dist} is normalized around the range of [0, 1], and the smaller, the better. 
See \autoref{sec:embodiment_bench_details} for the details.

\subsection{Behavior Distillation on \proposedbench}
\label{sec:distillation_results}
We systematically evaluate three types of morphology-task generalization; in-distribution, compositional, and out-of-distribution generalization through behavior distillation.
As baselines, we compare MLP, GNN~\citep{wang2018nervenet} with \proposedvariant{}, Transformer with \proposedvariant{} or \proposedshort{}, and tokenized \proposedshort{}, similar to \citet{reed2022gato} (see \autoref{sec:gato_representtion} for the details).

In in-distribution settings, we prepare 50 environments and 60 unimal environments adapted from \citet{gupta2022metamorph} (see Appendix~\ref{sec:unimal_details} for the details) for both training and evaluation.
Compositional and out-of-distribution settings evaluate zero-shot transfer.
In compositional settings, we test morphology and task generalization separately; we prepare 38 train environments and 12 test environments with hold-out morphologies for morphology evaluation, and leverage 50 train environments and prepare 9 test environments with unseen task for task evaluation.
In out-of-distribution settings, we also leverage 50 environments as a training set, and define 27 environments with diversified morphologies and unseen task as an evaluation set.
The proficient behavioral data contains 12k transitions per each environment.
See \autoref{sec:environment_division} for the details of environment division.

\autoref{tab:distillation} reveals that \proposedshort{} achieves the best multi-task goal-reaching performances among other possible combinations  in all the aspects of generalization.
Comparing average normalized distance, \proposedshort{} improves the multi-task performance against the second best, \proposedvariant{}, by 23\% in in-distribution evaluation.
Following previous works~\citep{kurin2020my, gupta2022metamorph}, Transformer with \proposedvariant{} achieves better goal-reaching behaviors than GNN. 
In compositional and out-of-distribution zero-shot evaluation, \proposedshort{} outperforms other choices by 14 $\sim$ 18\%.
Moreover, the compositional zero-shot performance of \proposedshort{} is comparable with the performance of \proposedvariant{} in in-distribution settings.
These results imply \proposedshort{} might be the better formulation to realize the morphology-task generalization.

\begin{figure}[t]
\centering
\includegraphics[width=0.85\linewidth]{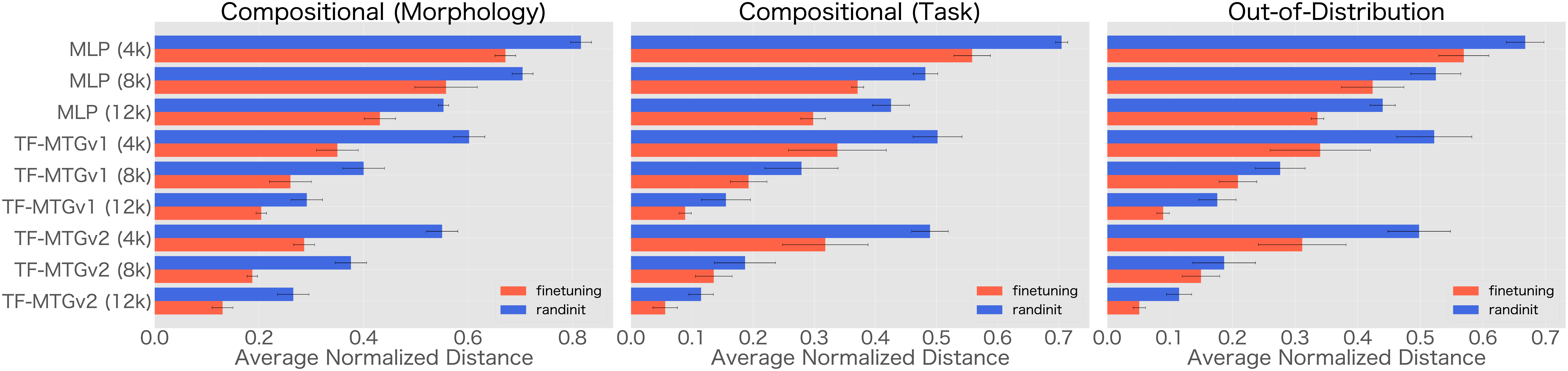}
\vskip -0.1in
\caption{Multi-task goal-reaching performances on fine-tuning (multi-task imitation) for compositional and out-of-distribution evaluation.
These results reveal that fine-tuning outperforms random initialization in all settings, and fine-tuned \proposedshort{} outperforms others by 50 $\sim$ 55 \%.
See \autoref{sec:additional_results} for the detailed scores.
}
\label{fig:finetuning_all}
\vskip -0.1in
\end{figure}

\subsection{Does \proposedREP{} Obtain Better Prior for Control?}
\label{sec:zero_shot_finetuning_results}

To reveal whether the distilled policy obtains reusable inductive bias for unseen morphology or task, we test the fine-tuning performance for multi-task imitation learning on \proposedbench.
We adopt the same morphology-task division for  compositional and out-of-distribution evaluation in Section~\ref{sec:distillation_results}.
\autoref{fig:finetuning_all} shows that fine-tuning outperforms random initialization in all settings, which suggests that behavior-distilled policy works as a better prior knowledge for control.
The same as zero-shot transfer results in Section~\ref{sec:distillation_results}, \proposedshort{} outperforms other baselines, and is better than the second best, \proposedvariant{} by 50 $\sim$ 55\%.
Furthermore, \proposedshort{} could work even with a small amount of dataset (4k, 8k); for instance, in compositional morphology evaluation (left in \autoref{fig:finetuning_all}), \proposedshort{} trained with 8k transitions still outperforms competitive combinations with 12k transitions, which indicates better sample efficiency (see \autoref{sec:additional_results} for the detailed scores).
These results suggest \proposedshort{} significantly captures the structure of \proposedrep{} as prior knowledge for downstream tasks.

\updates{
\subsection{What is the best node features for \proposedREP{}?}
\label{sec:node_feature_selection}
}

\updates{


In the agent's system, there are a lot of observable variables per module: such as Cartesian position (\textbf{p}), Cartesian velocity (\textbf{v)}, quaternion (\textbf{q}), angular velocity (\textbf{a}), joint angle (\textbf{ja}), joint range (\textbf{jr}), limb id (\textbf{id}), joint velocity (\textbf{jv}), relative position (\textbf{rp}), relative rotation (\textbf{rr}), and morphological information (\textbf{m}).
Morphological information contains module's shape, mass, inertia, actuator's gear, dof-index, etc.
To shed light on the importance of the local representation for generalization, we execute an extensive ablation of node feature selections and prepossessing.
In prior works, \citet{huang2020policy} and others~\citep{kurin2020my,hong2022structureaware} used \{\textbf{p}, \textbf{v}, \textbf{q}, \textbf{a}, \textbf{ja}, \textbf{jr}, \textbf{id}\} and \citet{gupta2022metamorph} used \{\textbf{p}, \textbf{v}, \textbf{q}, \textbf{a}, \textbf{ja}, \textbf{jr}, \textbf{jv}, \textbf{rp}, \textbf{rr}, \textbf{m}\}.
Considering the intersection of those, we define \{\textbf{p}, \textbf{v}, \textbf{q}, \textbf{a}, \textbf{ja}, \textbf{jr}\} as \texttt{base\_set} and test the combination to other observations (\textbf{jv}, \textbf{id}, \textbf{rp}, \textbf{rr}, \textbf{m}).

In the experiments, we evaluate in-distribution generalization, as done in Section~\ref{sec:distillation_results}, with \proposedshort{} representation.
\autoref{tab:node_feat} shows that, while some additional features (\textbf{id}, \textbf{rp}, \textbf{m}) contribute to improving the performance, the most effective feature can be morphological information (\textbf{m}), which suggests \texttt{base\_set} contains sufficient features for control, and raw morphological properties (\textbf{m}) serves better task specification than manually-encoded information such as limb id (\textbf{id}).
The key observations are; (1) morphological information is critical for morphology-task generalization, and (2) extra observation might disrupt the performance, such as relative rotation between parent and child node (\textbf{rr}).
Throughout the paper, we treat \texttt{base\_set-m} as node features for \proposedrep{}. 

}

\begin{table*}[ht]
\begin{center}
\begin{small}
\scalebox{0.8}{
\begin{tabular}{l|ccccc|l}
\toprule
Node features & +\textbf{jv} & +\textbf{id} & +\textbf{rp} & +\textbf{rr} & +\textbf{m} & \textbf{Average Dist.} \\
\midrule
\texttt{base\_set} & & & & & & 0.4330 $\pm$ 0.02 \\
\texttt{base\_set-id} & & \textcolor{green}{\CheckmarkBold} & & & & 0.4090 $\pm$ 0.02 \\
\texttt{base\_set-rp} & & & \textcolor{green}{\CheckmarkBold} & & & 0.3820 $\pm$ 0.01 \\
\texttt{base\_set-rr} & &  &  & \textcolor{green}{\CheckmarkBold} & & 0.4543 $\pm$ 0.01 \\
\texttt{base\_set-m} & &  &  &  & \textcolor{green}{\CheckmarkBold} & \textbf{0.3128 $\pm$ 0.02}\\
\texttt{base\_set-rp-rr} & &  & \textcolor{green}{\CheckmarkBold} & \textcolor{green}{\CheckmarkBold} & & 0.3869 $\pm$ 0.01 \\
\texttt{base\_set-jv-rp-rr} & \textcolor{green}{\CheckmarkBold} &  & \textcolor{green}{\CheckmarkBold} & \textcolor{green}{\CheckmarkBold} & & 0.4000 $\pm$ 0.01 \\
\texttt{base\_set-jv-rp-rr-m} & \textcolor{green}{\CheckmarkBold} &  & \textcolor{green}{\CheckmarkBold} & \textcolor{green}{\CheckmarkBold} & \textcolor{green}{\CheckmarkBold} & 0.3323 $\pm$ 0.01\\
\bottomrule
\end{tabular}
}
\end{small}
\end{center}
\vskip -0.1in
\caption{
\updates{
Offline node feature selection. We compare the combination of Cartesian position (\textbf{p}), Cartesian velocity (\textbf{v)}, quaternion (\textbf{q}), angular velocity (\textbf{a}), joint angle (\textbf{ja}), joint range (\textbf{jr}), limb id (\textbf{id}), joint velocity (\textbf{jv}), relative position (\textbf{rp}), relative rotation (\textbf{rr}), and morphological information (\textbf{m}).
\texttt{base\_set} is composed of \{\textbf{p}, \textbf{v}, \textbf{q}, \textbf{a}, \textbf{ja}, \textbf{jr}\}. 
These results suggest \texttt{base\_set} contains sufficient features for control, and the most effective feature seems morphological information (\textbf{m}) for better task specification.
}
}
\label{tab:node_feat}
\vskip -0.1in
\end{table*}




\section{Discussion and Limitation}

While the experimental evaluation on \proposedbench{} implies that \proposedrep{} is a simple and effective method to distill the diverse proficient behavioral data into a generalizable single policy, there are some limitations. For instance, we  focus on distillation from expert policies only, but it is still unclear whether \proposedrep{} works with moderate or random quality behaviors in offline RL~\citep{fujimoto2019off,levine2020offline,fujimoto2021minimalist}.
Combining distillation with iterative data collection~\citep{ghosh2021learning,matsushima2021deploy} or online fine-tuning~\citep{Li2022Phasic} would be a promising future work.
In addition, we avoided tasks where expert behaviors cannot be generated easily by single-task RL without fine-scale reward engineering or human demonstrations; incorporating such datasets or bootstrapping single-task RL from the distilled policy could be critical for scaling the pipeline to more complex tasks such as open-ended and dexterous manipulation~\citep{lynch2019learning,ghasemipour22blocks,bidex}.
Since \proposedrep{} only uses readily accessible features in any simulator and could be automatically defined through URDFs or MuJoCo XMLs, in future work we aim to keep training our best \proposedrep{} architecture policy on additional data from more functional and realistic control behaviors from other simulators like MuJoCo~\citep{yu2019meta,tassa2018deepmind}, PyBullet~\citep{shridhar2022cliport}, IsaacGym~\citep{bidex,peng2021amp}, and Unity~\citep{juliani2018unity}, and show it achieves better scaling laws than other representations~\citep{reed2022gato} on broader morphology-task families.

\section{Conclusion}
The broader range of behavior generalization is a promising paradigm for RL.
To achieve morphology-task generalization, we propose \proposedrep{}, which expresses the agent's modular observations, actions, and goals as a unified graph representation while preserving the geometric task structure.
As a test bed for morphology-task generalization, we also develop \proposedbench{}, which enables the scalable procedural generation of agents and tasks with minimal blueprints.
Fast-generated behavior datasets of \proposedbench{} with RL allow efficient representation and architecture selection through supervised learning, and \proposedshort{}, variant of \proposedrep{}, achieves the best multi-task performances among other possible designs (MLP, GNN and Transformer with \proposedvariant{}, and tokenized-\proposedshort{}, etc), outperforming them in in-distribution evaluation (by 23 \%), zero-shot transfer among compositional or out-of-distribution evaluation (by 14 $\sim$ 18 \%) and fine-tuning for downstream multi-task imitation~(by 50 $\sim$ 55 \%).
We hope our work will encourage the community to explore scalable yet incremental approaches for building a universal controller.

\clearpage
\subsubsection*{Acknowledgments}
This work was supported by JSPS KAKENHI Grant Number JP22J21582.
We thank Mitsuhiko Nakamoto and Daniel C. Freeman for converting several MuJoCo agents to Brax, So Kuroki for the support on implementations, Yujin Tang, Kamyar Ghasemipour, Yingtao Tian, and Bert Chan for helpful feedback on this work.

\bibliography{reference}
\bibliographystyle{iclr2023_conference}

\clearpage
\appendix
\section*{Appendix}
\section{Details of Implementation}
\label{sec:implementation}

The hyperparameters we used are listed in \autoref{tab:hyperparam}. 
We implement MLP and Transformer policy with Jax~\citep{jax2018github} and Flax~\citep{flax2020github}.
For the implementation of GNN policy, we use a graph neural network library, Jraph~\citep{jraph2020github}.

\begin{table*}[ht]
\begin{center}
\begin{small}
\scalebox{1.}{
\begin{tabular}{lll}
\toprule
\textbf{Method} & \textbf{Hyperparameter} & \textbf{Value} \\
\midrule
Shared & Learning rate & 3e-4 \\
 & Batch size & 256 \\
 & Gradient clipping & 0.1 \\
 & Activation function & ReLU \\
 & Gradient steps & 100k \\
\midrule
MLP & Hidden size & 1024 \\
 & \# of layers & 2 \\
\midrule
Transformer & Embedding size & 256 \\
 & Attention hidden size & 512 \\
 & \# of attention heads & 2 \\
 & \# of attention layers & 3 \\
\midrule
GNN & Hidden size & 256 \\
 & \# of layers (per node) & 3 \\
\bottomrule
\end{tabular}
}
\end{small}
\end{center}
\caption{Hyperparameters for each method.}
\label{tab:hyperparam}
\end{table*}

\updates{
\textbf{Transformer for \proposedREP{}}~
Transformer first encodes \proposedrep{} to the latent representation vector $\bm{z_0}$ with shared single-layer MLP and learnable position embedding (PE) (in case of \proposedvariant{}, we omit $s_g$, but include it  to corresponding node observation $s^{i}$ instead):
\begin{equation}
    \bm{z_0} = [\text{MLP}(s^{1}), \dots, \text{MLP}(s^{|\bm{V_m}|}), ~\text{MLP}(s_g)] + ~\text{PE},  \nonumber
\end{equation}
then multi-head attention (MHA)~\citep{vaswani2017attention} and layer normalization (LayerNorm)~\citep{ba2016layer} are recursively applied to latent representation $\bm{z_l}$ at $l$-th layer,
\begin{align}
    \bm{z'_l} &= \text{LayerNorm}(\text{MHA}(\bm{z_{l-1}}) + \bm{z_{l-1}}),  \nonumber \\
    \bm{z_l} &= \text{LayerNorm}(\text{MLP}(\bm{z'_l}) + \bm{z'_l}), \qquad\qquad\qquad (l = 1, ..., L).  \nonumber
\end{align}
Before decoding the action per module from the last-layer latent representation $\bm{z_L}$, we employ the residual connection of the node features~\citep{kurin2020my},
\begin{equation}
    \bm{a^m} = [\text{MLP}([z_L^1, s^{1}]), \dots, ~\text{MLP}([z_L^{|\bm{V_m}|}, s^{|\bm{V_m}|}]), ~\text{MLP}([z_L^{|\bm{V_m}|+1}, s_{g}])],  \nonumber
\end{equation}
where shared MLP has a single layer and tanh activation to clip the output within the range of [-1, 1]. We mask out the output from the goal modules or modules that have no actuators to ensure action size as $|\bm{E}_m|$.

\textbf{Details of \proposedREP{} v1 and v2}~
Our proposed \proposedrep{} expresses the agent's observations, actions, and goals or tasks in a unified graph representation, while preserving the geometric structure of the task.
\proposedRep{} v1 accepts the morphological graph, encoded from the agent's geometric information, as an input-output interface, and merges positional goal information as a part of corresponding node features, and \proposedrep{} v2 treats given goals as extra disjoint nodes of morphological graph.
In practice, we pad the incompatible parts of goal information by 0. For both \proposedvariant{} and \proposedshort{}, we prepare the binary vectors that have indicators to the target (or goal) nodes. \proposedvariant{} obtains the products of the goal information and binary vectors, then leverages them as extra node features. \proposedshort{} just takes such binary vectors as extra node features (and also includes goals as extra nodes).

}

\clearpage
\section{Details of \proposedbench}
\label{sec:embodiment_bench_details}

\subsection{Morphology}

\begin{figure*}[ht]
    \centering
    \includegraphics[width=1\linewidth]{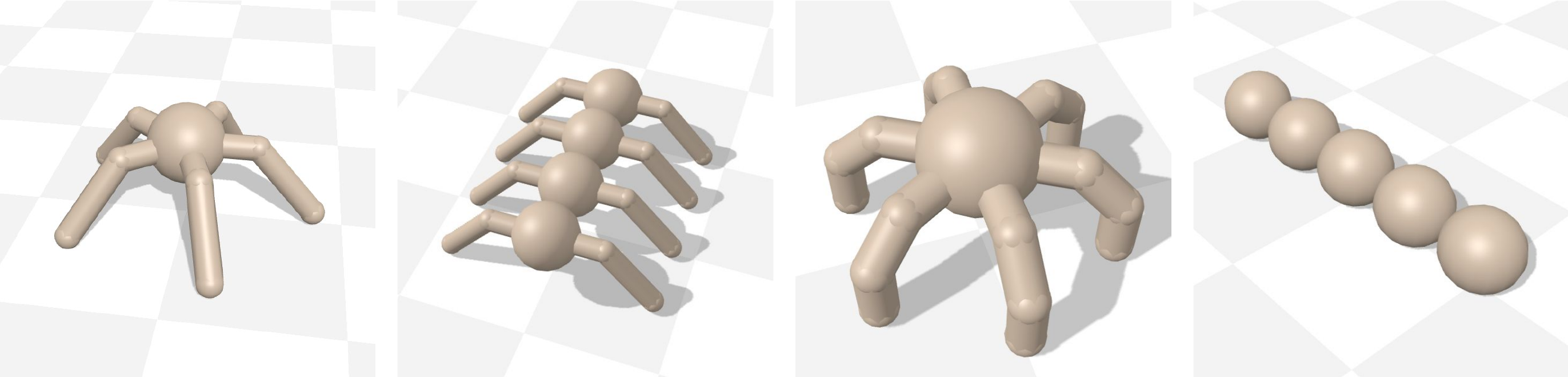}
    \caption{
    Examples of procedurally-generated morphology from base blueprints in \proposedbench{}.
    From left to right, each figure shows the example of 5-leg ant, 4-body centipede, 6-leg claw, and 5-body worm.
    }
    \label{fig:morphology_example}
\end{figure*}

We prepare 4 base blueprints; ant, centipede, claw, and worm, for procedural scene generation~(\autoref{fig:morphology_example}).
Base ant has 1 torso and 2 legs with 2 joints per limb, and we could procedurally generate the agents with different number of limbs.
Base centipede has 2 bodies and 4 legs with 2 joints per limb, and we could procedurally generate the agents with different number of bodies.
Base claw has 1 torso and 2 legs with 4 joints per limb (each leg consists of 3 modules), and we could procedurally generate the agents with different number of limbs.
Base worm has 2 bodies and no legs, and we could procedurally generate the agents with different number of bodies.

\begin{figure*}[ht]
    \centering
    \includegraphics[width=1\linewidth]{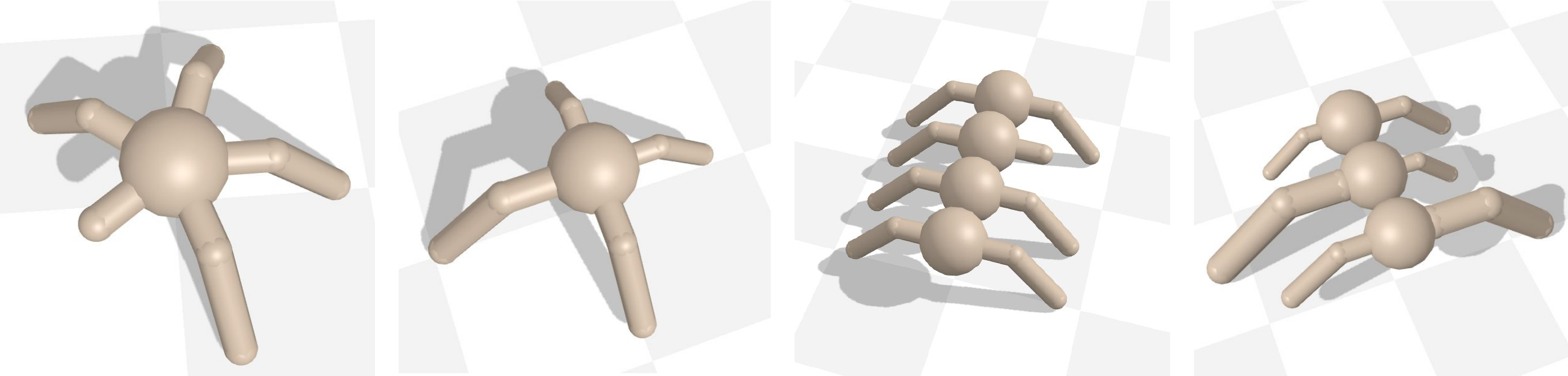}
    \caption{Examples of diversified morphology from base blueprints in \proposedbench{}.
    From left to right, each figure shows the example of 5-leg-1-missing ant, 5-leg-size-randomized ant, 4-body-1-missing centipede, and 3-body-size-randomized centipede.
    }
    \label{fig:morphology_diverse_example}
\end{figure*}

Furthermore, we develop the functionality for morphology diversification, with missing, mass, and size parameters~(\autoref{fig:morphology_diverse_example}).
Missing randomization lacks one module at one leg. This might be an equivalent situation that one leg is broken.
Mass randomization changes the default mass of each module with specified scales.
The appearance does not change, but certainly, the dynamics would differ.
Size randomization changes the default length and radius of each module with specified scales.

\subsection{Task}
\label{sec:mxt_bench_task}

\begin{figure*}[ht]
    \centering
    \includegraphics[width=1\linewidth]{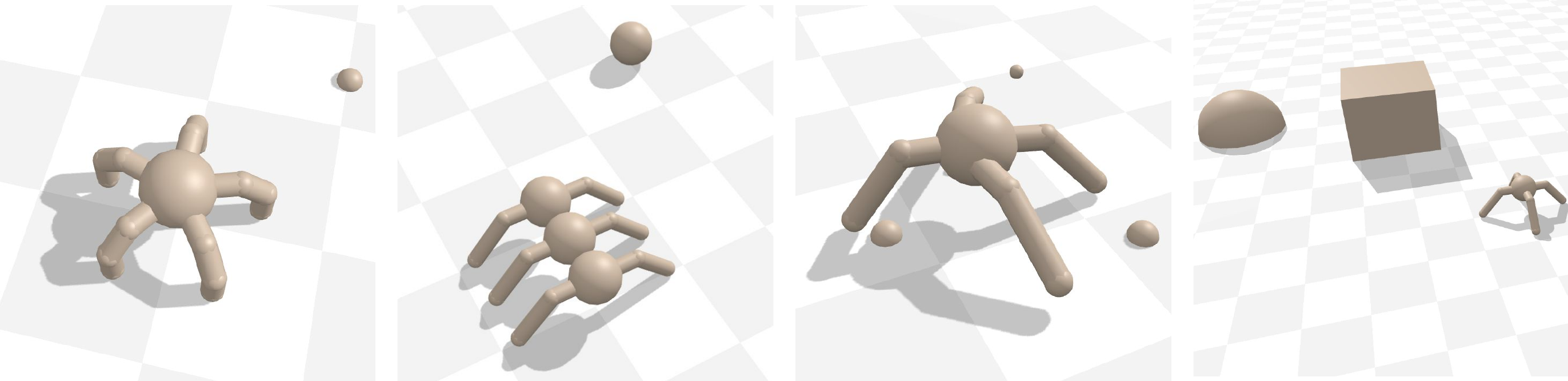}
    \caption{
    Examples of pre-defined task in \proposedbench{}.
    From left to right, each figure shows the example of reach, touch, twisters, and push task.
    }
    \label{fig:task_example}
\end{figure*}

We prepare 4 base tasks with parameterized goal distributions; reach, touch, twisters, and push, for procedural task generation~(\autoref{fig:task_example}).
Reach task requires the agents to put their one leg to the given goal position (XY). The variant, reach\_hard task, represents that the goal distribution is farther than reach task.
Touch task requires the agents to contact their body or torso to the movable ball (i.e. movable ball is a goal).
Twisters is a multi-goal problem; the agents should satisfy given goals at the same time.
There are two basic constraints; reach and handsup.
Handsup requires the agents to raise their one leg to the given goal Z height.
Twisters has some combinations like reach\_handsup, reach\_hard\_handsup, reach2\_handsup, or reach\_handsup2. For instance, in reach\_handsup2, the agents should put their one leg to the given goal position, and raise their two legs to the given goal heights simultaneously.
\updates{The XY position goals are sampled from uniform distribution with a predefined donut-shape range, and Z height goals are also from uniform distribution with predefined range.}
Push task requires the agents to move the box object to the given goal position (XY) \updates{sampled from uniform distribution with predefined range}. 
Since it has richer interaction with the object, push might be a more difficult task than those three.
We use this task in \autoref{sec:task_ood_push}.

\subsection{Custom Morphology}
\label{sec:unimal_details}

\begin{figure*}[ht]
    \centering
    \includegraphics[width=1\linewidth]{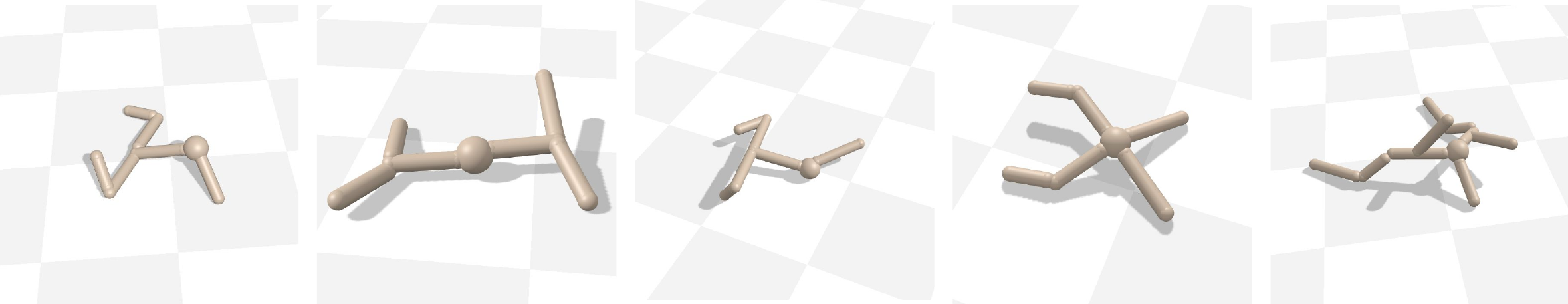}
    \caption{Examples of unimal agents, adapted from \citet{gupta2022metamorph}}
    \label{fig:unimal_example}
\end{figure*}

\proposedbench{} also supports custom morphology import used in previous work.
For instance, \citet{gupta2022metamorph} propose unimal agents that are generated via evolutional strategy and designed for MuJoCo.
Since they are not manually designed, their morphologies seem more diverse than our ant, centipede, claw, and worm.
We inspect unimals whether they are suitable for goal-reaching, and include 72 morphologies from there (and use 60 morphologies for the experiments).
\autoref{fig:unimal_example} shows some example of unimals.

\subsection{Comparison to Existing Benchmark}
\label{sec:existing_bench}
Since the current RL community has not paid much attention to embodied control so far,
there are no suitable benchmarks to quantify the generalization over various tasks beyond single locomotion tasks and morphologies at the same time. 
In addition, the scalability to various morphologies or tasks seems to be required for the benchmark, because we should avoid ``overfitting'' to manually-designed tasks. 

As summarized in \autoref{fig:embodiment_bench}, MuJoCo~\citep{todorov2012mujoco} or DM Control~\citep{tassa2018deepmind}, the most popular benchmarks in the continuous control domain, could not evaluate task or morphology generalization; they only have a single morphology for a single task as a pre-defined environment.
\citet{yu2019meta} propose a robot manipulation benchmark for meta RL, but it does not care about the morphology. Furthermore, while it has quite a diverse set of tasks, the scalability of environments seems to be limited.
In contrast, previous morphology-agnostic RL works~\citep{huang2020policy,kurin2020my,hong2022structureaware,br2022anymorph} have a set of different morphologies adapted from MuJoCo agents. \citet{gupta2022metamorph} also provide a much larger set of agents that are produced via joint-optimization of morphology and task rewards by the evolutionary strategy~\citep{gupta2021embodied}, with kinematics and dynamics randomization. However, those works only aim to solve single locomotion tasks, i.e. running forward as fast as possible.

\clearpage

\section{Details of Expert Data Generation}
\label{sec:data_gen_details}
\begin{figure*}[ht]
    \centering
    \includegraphics[width=1\linewidth]{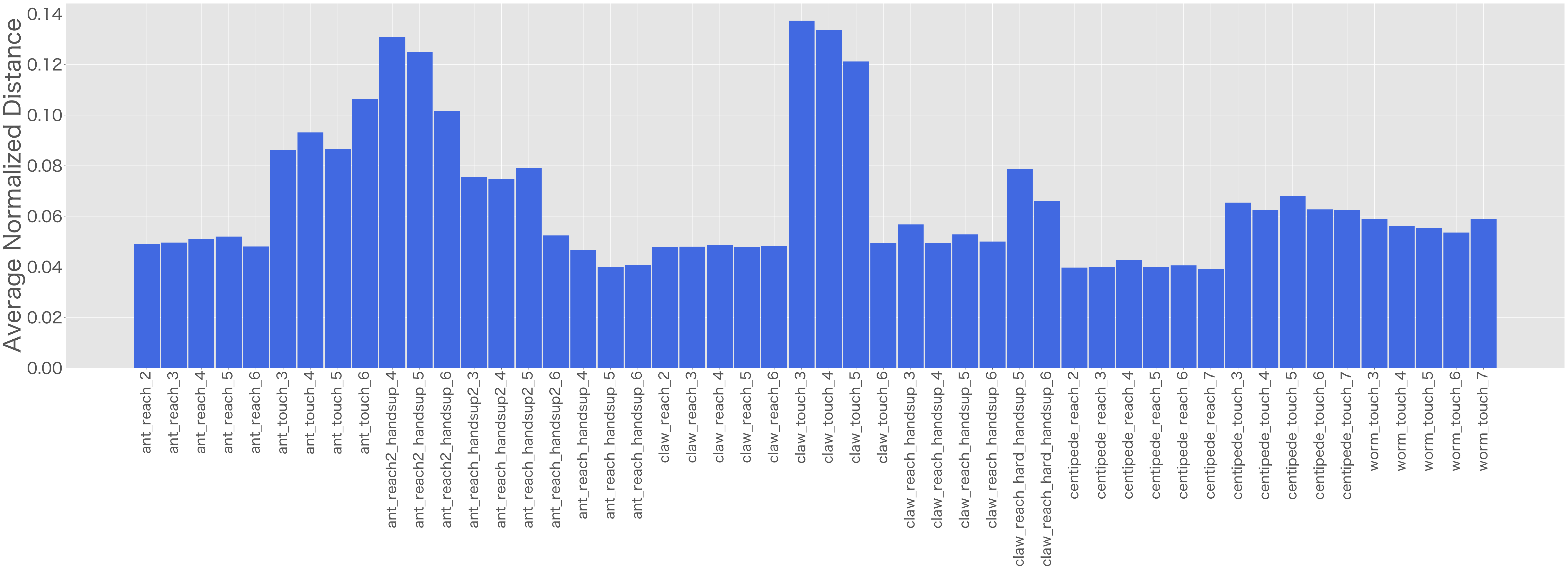}
    \caption{\updates{Average normalized final distance in each dataset (\autoref{tab:zs_morph_entry}).}}
    \label{fig:dataset}
\end{figure*}

We train the single-task PPO~\citep{Schulman2017PPO} at each environment to obtain the expert policy.
For the reward function, we basically adopt dense reward (except for push task) $-d_{\psi}(\bm{s}^m, s_g)$ until agents satisfy a given condition, $\mathbbm{1}[d_{\psi}(\bm{s}^m, s_g) \leq d^{m,\psi}_{\min}]$. See \autoref{sec:environment_division} for the threshold $d^{m,\psi}_{\min}$.
After convergence, we collect the proficient behaviors.
Unless otherwise specified, we use 12k transitions per environment.
The average normalized final distances of those datasets are mostly less than 0.1~\updates{(\autoref{fig:dataset})}.

\section{Environments Division}
\label{sec:environment_division}
Throughout the experiments, we test a lot of morphology-task combinations to investigate the in-distribution generalization, compositional generalization for morphology and task, and out-of-distribution generalization.
In this section, we list up the combination of environments used in the experiments.

\autoref{tab:zs_morph_entry} explains the combinations for the experiments of in-distribution generalization, and compositional generalization for morphology (both zero-shot transfer and fine-tuning) in \autoref{tab:distillation} and \autoref{fig:finetuning_all}.
For compositional morphology evaluation, we use Morph-Train division as training dataset and Morph-Test division as evaluation environments.
For the evaluation of dataset size and the number of
morphology-task combinations~(\autoref{sec:data_entry}), we use In-Distribution, 12 Env, and 25 Env division as train datasets and test environments.
\updates{Following prior works~\citep{huang2020policy,kurin2020my,gupta2022metamorph}, we have chosen Morph-Test division that would not exceed the maximum number of nodes in Morph-Train division for convenience.
We have checked our trained model could be generalized to the agents with an extra number of limbs (e.g. 7-leg ant or 8-body centipede), but have not extensively increased the number of limbs until the model could not deal with them.
We note that this may limit the morphology generalization in the open-ended evaluation.}

\autoref{tab:unimal_id} also explains the combinations for the experiments of in-distribution generalization with unimals~\citep{gupta2022metamorph} in \autoref{tab:distillation} and \autoref{tab:unimal}.

\autoref{tab:zs_task_ood_entry} provides the combinations for the experiments of compositional generalization for task and out-of-distribution generalization (both zero-shot transfer and fine-tuning) in \autoref{tab:distillation} and \autoref{fig:finetuning_all}.
We use them as test environments, and for training datasets, we leverage In-Distribution division in \autoref{tab:zs_morph_entry}.

In addition, we extensively evaluate the compositional generalization for task and out-of-distribution generalization with more different unseen task, such as push~(see~\autoref{sec:task_ood_push} for the details).
\autoref{tab:zs_task_ood_entry_push} also shows the environment division for both zero-shot transfer and fine-tuning.

\begin{table*}[ht]
\begin{center}
\begin{small}
\scalebox{0.65}{
\begin{tabular}{llccccccc}
\toprule
\textbf{Sub-domain} & \textbf{Environment} & $d_{\min}^{\psi}$ & $d_{\max}^{\psi}$ & \textbf{In-Distribution} & \textbf{Morph-Train} & \textbf{Morph-Test} & \textbf{12 Env} & \textbf{25 Env} \\
\midrule
\multirow{5}{*}{ant\_reach} & ant\_reach\_2 & 0.1 & 8.75 & \CheckmarkBold & \CheckmarkBold &  \\ 
& ant\_reach\_3 & 0.1 & 8.75 & \CheckmarkBold & \CheckmarkBold &  \\ 
& ant\_reach\_4 & 0.1 & 8.75 & \CheckmarkBold &  & \CheckmarkBold & \CheckmarkBold & \CheckmarkBold \\ 
& ant\_reach\_5 & 0.1 & 8.75 & \CheckmarkBold & \CheckmarkBold &  \\ 
& ant\_reach\_6 & 0.1 & 8.75 & \CheckmarkBold & \CheckmarkBold &  &  & \CheckmarkBold \\ 
\midrule
\multirow{4}{*}{ant\_touch} & ant\_touch\_3 & 0.5 & 3.5 & \CheckmarkBold & \CheckmarkBold &  &  & \CheckmarkBold \\ 
& ant\_touch\_4 & 0.5 & 3.5 & \CheckmarkBold &  & \CheckmarkBold & \CheckmarkBold & \CheckmarkBold \\
& ant\_touch\_5 & 0.5 & 3.5 & \CheckmarkBold & \CheckmarkBold &  \\ 
& ant\_touch\_6 & 0.5 & 3.5 & \CheckmarkBold & \CheckmarkBold  &  &  & \CheckmarkBold \\ 
\midrule
\multirow{10}{*}{ant\_twisters} & ant\_reach2\_handsup\_4 & \{0.1, 0.1, 0.1\} & \{1.4, 1.4, 0.625\} & \CheckmarkBold &  & \CheckmarkBold \\ 
& ant\_reach2\_handsup\_5 & \{0.1, 0.1, 0.1\} & \{1.4, 1.4, 0.625\} & \CheckmarkBold & \CheckmarkBold &  & \CheckmarkBold & \CheckmarkBold \\ 
& ant\_reach2\_handsup\_6 & \{0.1, 0.1, 0.1\} & \{1.4, 1.4, 0.625\} & \CheckmarkBold & \CheckmarkBold & & & \CheckmarkBold \\ 
& ant\_reach\_handsup2\_3 & \{0.1, 0.1, 0.1\} & \{1.5, 0.55, 0.55\} & \CheckmarkBold & \CheckmarkBold  &  \\ 
& ant\_reach\_handsup2\_4 & \{0.1, 0.1, 0.1\} & \{1.5, 0.55, 0.55\} & \CheckmarkBold &  & \CheckmarkBold & & \CheckmarkBold \\ 
& ant\_reach\_handsup2\_5 & \{0.1, 0.1, 0.1\} & \{1.5, 0.55, 0.55\} & \CheckmarkBold & \CheckmarkBold &  & \CheckmarkBold & \CheckmarkBold \\ 
& ant\_reach\_handsup2\_6 & \{0.1, 0.1, 0.1\} & \{1.5, 0.55, 0.55\} & \CheckmarkBold & \CheckmarkBold &  \\ 
& ant\_reach\_handsup\_4 & \{0.1, 0.1\} & \{4.5, 0.55\} & \CheckmarkBold &  & \CheckmarkBold & & \CheckmarkBold \\ 
& ant\_reach\_handsup\_5 & \{0.1, 0.1\} & \{4.5, 0.55\} & \CheckmarkBold & \CheckmarkBold &  & \CheckmarkBold & \CheckmarkBold \\ 
& ant\_reach\_handsup\_6 & \{0.1, 0.1\} & \{4.5, 0.55\} & \CheckmarkBold & \CheckmarkBold &  \\ 
\midrule
\multirow{5}{*}{claw\_reach} & claw\_reach\_2 & 0.1 & 8.75 & \CheckmarkBold & \CheckmarkBold &  &  & \CheckmarkBold \\
& claw\_reach\_3 & 0.1 & 8.75 & \CheckmarkBold & \CheckmarkBold &  \\ 
& claw\_reach\_4 & 0.1 & 8.75 & \CheckmarkBold & \CheckmarkBold & & \CheckmarkBold & \\ 
& claw\_reach\_5 & 0.1 & 8.75 & \CheckmarkBold &  & \CheckmarkBold &  & \CheckmarkBold \\ 
& claw\_reach\_6 & 0.1 & 8.75 & \CheckmarkBold & \CheckmarkBold &  \\ 
\midrule
\multirow{4}{*}{claw\_touch} & claw\_touch\_3 & 0.5 & 3.5 & \CheckmarkBold & \CheckmarkBold &  \\ 
& claw\_touch\_4 & 0.5 & 3.5 & \CheckmarkBold & \CheckmarkBold  &  & \CheckmarkBold & \CheckmarkBold \\
& claw\_touch\_5 & 0.5 & 3.5 & \CheckmarkBold &  & \CheckmarkBold \\ 
& claw\_touch\_6 & 0.5 & 3.5 & \CheckmarkBold & \CheckmarkBold  &  &  & \CheckmarkBold \\ 
\midrule
\multirow{6}{*}{claw\_twisters} & claw\_reach\_handsup\_3 & \{0.1, 0.1\} & \{1.4, 0.625\} & \CheckmarkBold & \CheckmarkBold & & & \CheckmarkBold \\ 
& claw\_reach\_handsup\_4 & \{0.1, 0.1\} & \{1.4, 0.625\} & \CheckmarkBold & \CheckmarkBold &  \\ 
& claw\_reach\_handsup\_5 & \{0.1, 0.1\} & \{1.4, 0.625\} & \CheckmarkBold &  & \CheckmarkBold & \CheckmarkBold & \CheckmarkBold \\ 
& claw\_reach\_handsup\_6 & \{0.1, 0.1\} & \{1.4, 0.625\} & \CheckmarkBold & \CheckmarkBold &  \\ 
& claw\_reach\_hard\_handsup\_5 & \{0.1, 0.1\} & \{4.5, 0.55\} & \CheckmarkBold &  & \CheckmarkBold & \CheckmarkBold & \CheckmarkBold \\ 
& claw\_reach\_hard\_handsup\_6 & \{0.1, 0.1\} & \{4.5, 0.55\} & \CheckmarkBold & \CheckmarkBold &  \\ 
\midrule
\multirow{6}{*}{centipede\_reach} & centipede\_reach\_2 & 0.1 & 3.0 & \CheckmarkBold & \CheckmarkBold &  \\ 
& centipede\_reach\_3 & 0.1 & 3.0 & \CheckmarkBold & \CheckmarkBold &  & \CheckmarkBold & \CheckmarkBold \\ 
& centipede\_reach\_4 & 0.1 & 3.0 & \CheckmarkBold & \CheckmarkBold &  \\ 
& centipede\_reach\_5 & 0.1 & 3.0 & \CheckmarkBold & \CheckmarkBold &  \\ 
& centipede\_reach\_6 & 0.1 & 3.0 & \CheckmarkBold &  & \CheckmarkBold \\ 
& centipede\_reach\_7 & 0.1 & 3.0 & \CheckmarkBold & \CheckmarkBold &  & & \CheckmarkBold \\ 
\midrule
\multirow{5}{*}{centipede\_touch} & centipede\_touch\_3 & 0.5 & 10.5 & \CheckmarkBold & \CheckmarkBold \\ 
& centipede\_touch\_4 & 0.5 & 10.5 & \CheckmarkBold & \CheckmarkBold &  & & \CheckmarkBold \\ 
& centipede\_touch\_5 & 0.5 & 10.5 & \CheckmarkBold & \CheckmarkBold &  & & \CheckmarkBold\\ 
& centipede\_touch\_6 & 0.5 & 10.5 & \CheckmarkBold &  & \CheckmarkBold & \CheckmarkBold  & \CheckmarkBold \\ 
& centipede\_touch\_7 & 0.5 & 10.5 & \CheckmarkBold & \CheckmarkBold &  \\ 
\midrule
\multirow{5}{*}{worm\_touch} & worm\_touch\_3 & 0.5 & 3.5 & \CheckmarkBold & \CheckmarkBold &  \\ 
& worm\_touch\_4 & 0.5 & 3.5 & \CheckmarkBold & \CheckmarkBold & & & \CheckmarkBold \\  
& worm\_touch\_5 & 0.5 & 3.5 & \CheckmarkBold &  & \CheckmarkBold & \CheckmarkBold &  \\ 
& worm\_touch\_6 & 0.5 & 3.5 & \CheckmarkBold & \CheckmarkBold & & & \CheckmarkBold \\  
& worm\_touch\_7 & 0.5 & 3.5 & \CheckmarkBold & \CheckmarkBold &  \\ 
\bottomrule
\end{tabular}
}
\end{small}
\end{center}
\caption{The combinations of environments used in the experiments of in-distribution generalization, and compositional generalization for morphology.}
\label{tab:zs_morph_entry}
\end{table*}

\begin{table*}[ht]
\begin{center}
\begin{small}
\scalebox{0.85}{
\begin{tabular}{llcccc}
\toprule
\textbf{Sub-domain} & \textbf{Environment} & $d_{\min}^{\psi}$ & $d_{\max}^{\psi}$ & \textbf{Task-Test} & \textbf{OOD-Test} \\
\midrule
\multirow{4}{*}{ant\_reach\_hard} & ant\_reach\_hard\_3 & 0.1 & 11.0 & \CheckmarkBold &  \\ 
& ant\_reach\_hard\_4 & 0.1 & 11.0 & \CheckmarkBold &  \\ 
& ant\_reach\_hard\_5 & 0.1 & 11.0 & \CheckmarkBold &  \\ 
& ant\_reach\_hard\_6 & 0.1 & 11.0 & \CheckmarkBold &  \\ 
\midrule
\multirow{6}{*}{centipede\_reach\_hard} & centipede\_reach\_hard\_3 & 0.1 & 5.5 & \CheckmarkBold &  \\ 
& centipede\_reach\_hard\_4 & 0.1 & 5.5 & \CheckmarkBold &  \\ 
& centipede\_reach\_hard\_5 & 0.1 & 5.5 & \CheckmarkBold &  \\ 
& centipede\_reach\_hard\_6 & 0.1 & 5.5 & \CheckmarkBold &  \\ 
& centipede\_reach\_hard\_7 & 0.1 & 5.5 & \CheckmarkBold &  \\ 
\midrule
\multirow{11}{*}{ant\_reach\_hard\_diverse} & ant\_reach\_hard\_4\_b & 0.1 & 11.0 & &  \CheckmarkBold \\ 
& ant\_reach\_hard\_5\_b & 0.1 & 11.0 & &  \CheckmarkBold \\ 
& ant\_reach\_hard\_4\_mass\_0.5\_1.0\_3.0 & 0.1 & 11.0 & &  \CheckmarkBold \\ 
& ant\_reach\_hard\_4\_mass\_0.5\_1.0\_1.0 & 0.1 & 11.0 & &  \CheckmarkBold \\ 
& ant\_reach\_hard\_4\_mass\_1.0\_3.0\_3.0 & 0.1 & 11.0 & &  \CheckmarkBold \\ 
& ant\_reach\_hard\_4\_size\_0.9\_1.0\_1.1 & 0.1 & 11.0 & &  \CheckmarkBold \\ 
& ant\_reach\_hard\_5\_mass\_0.5\_1.0\_3.0 & 0.1 & 11.0 & &  \CheckmarkBold \\ 
& ant\_reach\_hard\_5\_mass\_0.5\_1.0\_1.0 & 0.1 & 11.0 & &  \CheckmarkBold \\ 
& ant\_reach\_hard\_5\_mass\_1.0\_3.0\_3.0 & 0.1 & 11.0 & &  \CheckmarkBold \\ 
& ant\_reach\_hard\_5\_size\_0.9\_1.0\_1.1 & 0.1 & 11.0 & &  \CheckmarkBold \\ 
\midrule
\multirow{17}{*}{centipede\_reach\_hard\_diverse} & centipede\_reach\_hard\_3\_b\_r\_0 & 0.1 & 5.5 & & \CheckmarkBold \\
& centipede\_reach\_hard\_3\_b\_l\_0 & 0.1 & 5.5 & &  \CheckmarkBold \\ 
& centipede\_reach\_hard\_3\_b\_r\_1 & 0.1 & 5.5 & &  \CheckmarkBold \\ 
& centipede\_reach\_hard\_3\_b\_l\_1 & 0.1 & 5.5 & &  \CheckmarkBold \\ 
& centipede\_reach\_hard\_4\_b\_r\_0 & 0.1 & 5.5 & &  \CheckmarkBold \\ 
& centipede\_reach\_hard\_4\_b\_r\_1 & 0.1 & 5.5 & &  \CheckmarkBold \\ 
& centipede\_reach\_hard\_4\_b\_l\_1 & 0.1 & 5.5 & &  \CheckmarkBold \\ 
& centipede\_reach\_hard\_4\_b\_r\_2 & 0.1 & 5.5 & &  \CheckmarkBold \\ 
& centipede\_reach\_hard\_4\_b\_l\_2 & 0.1 & 5.5 & &  \CheckmarkBold \\ 
& centipede\_reach\_hard\_3\_size\_0.9\_1.0\_1.1 & 0.1 & 5.5 & &  \CheckmarkBold \\ 
& centipede\_reach\_hard\_3\_mass\_0.5\_1.0\_3.0 & 0.1 & 5.5 & &  \CheckmarkBold \\ 
& centipede\_reach\_hard\_3\_mass\_0.5\_1.0\_1.0 & 0.1 & 5.5 & &  \CheckmarkBold \\ 
& centipede\_reach\_hard\_3\_mass\_1.0\_3.0\_3.0 & 0.1 & 5.5 & &  \CheckmarkBold \\ 
& centipede\_reach\_hard\_4\_size\_0.9\_1.0\_1.1 & 0.1 & 5.5 & &  \CheckmarkBold \\ 
& centipede\_reach\_hard\_4\_mass\_0.5\_1.0\_3.0 & 0.1 & 5.5 & &  \CheckmarkBold \\ 
& centipede\_reach\_hard\_4\_mass\_0.5\_1.0\_1.0 & 0.1 & 5.5 & &  \CheckmarkBold \\
& centipede\_reach\_hard\_4\_mass\_1.0\_3.0\_3.0 & 0.1 & 5.5 & &  \CheckmarkBold \\
\bottomrule
\end{tabular}
}
\end{small}
\end{center}
\caption{The combinations of environments used in the experiments of compositional generalization for task and out-of-distribution generalization.}
\label{tab:zs_task_ood_entry}
\end{table*}

\begin{table*}[t]
\begin{center}
\begin{small}
\scalebox{0.7}{
\begin{tabular}{l|cc|ccc}
\toprule
\textbf{unimal\_id} & $d_{\min}^{\psi}$ & $d_{\max}^{\psi}$ & \textbf{reach} & \textbf{touch} & \textbf{twisters}  \\
\midrule
5506-0-13-17-12-26-41 &  &  &  \\
5506-12-12-01-11-30-00 & 0.1 & 8.75 & \CheckmarkBold &  &  \\
5506-15-16-01-14-17-18 & 0.35 & 3.5 &  & \CheckmarkBold &  \\
5506-6-5-01-14-20-42 & \{0.1, 0.1\} & \{1.5, 0.55\} &  &  & \CheckmarkBold \\
5506-0-2-01-15-36-43 & \{0.1, 0.1\} & \{1.5, 0.55\} &  &  & \CheckmarkBold \\
5506-12-12-01-15-33-01 & 0.1 & 8.75 & \CheckmarkBold &  &  \\
5506-15-16-02-22-21-06 & 0.1 & 8.75 & \CheckmarkBold &  &  \\
5506-6-8-17-09-59-06 &  &  &  \\
5506-0-5-01-12-45-36 & \{0.1, 0.1\} & \{1.5, 0.55\} &  &  & \CheckmarkBold \\
5506-12-14-01-15-22-01 & 0.1 & 8.75 & \CheckmarkBold &  &  \\
5506-2-0-01-11-27-44 & \{0.1, 0.1\} & \{1.5, 0.55\} &  &  & \CheckmarkBold \\
5506-7-6-17-12-20-01 & 0.35 & 3.5 &  & \CheckmarkBold &  \\
5506-0-7-01-15-34-13 & 0.35 & 3.5 &  & \CheckmarkBold &  \\
5506-12-6-17-08-36-18 &  &  &  \\
5506-2-16-01-10-58-23 & 0.1 & 8.75 & \CheckmarkBold &  &  \\
5506-8-11-01-15-28-53 & 0.35 & 3.5 &  & \CheckmarkBold &  \\
5506-1-12-17-11-10-12 & 0.1 & 8.75 & \CheckmarkBold &  &  \\
5506-12-6-17-12-20-06 & 0.1 & 8.75 & \CheckmarkBold &  &  \\
5506-2-17-17-10-16-02 & 0.1 & 8.75 & \CheckmarkBold &  &  \\
5506-8-12-01-13-32-46 & 0.35 & 3.5 &  & \CheckmarkBold &  \\
5506-1-13-17-12-03-16 & 0.1 & 8.75 & \CheckmarkBold &  &  \\
5506-12-8-01-14-50-41 & \{0.1, 0.1\} & \{1.5, 0.55\} &  &  & \CheckmarkBold \\
5506-2-9-17-11-10-42 &  &  &  \\
5506-8-16-01-13-07-43 & 0.1 & 8.75 & \CheckmarkBold &  &  \\
5506-1-15-17-07-32-47 & \{0.1, 0.1\} & \{1.5, 0.55\} &  &  & \CheckmarkBold \\
5506-13-10-17-12-25-45 & 0.1 & 8.75 & \CheckmarkBold &  &  \\
5506-3-10-01-14-19-06 & 0.1 & 8.75 & \CheckmarkBold &  &  \\
5506-8-16-02-14-47-12 & \{0.1, 0.1\} & \{1.5, 0.55\} &  &  & \CheckmarkBold \\
5506-1-2-02-20-28-11 & \{0.1, 0.1\} & \{1.5, 0.55\} &  &  & \CheckmarkBold \\
5506-13-17-01-16-09-18 & \{0.1, 0.1\} & \{1.5, 0.55\} &  &  & \CheckmarkBold \\
5506-3-15-01-14-36-50 & 0.35 & 3.5 &  & \CheckmarkBold &  \\
5506-8-17-17-09-38-29 &  &  &  \\
5506-1-5-02-19-23-33 & \{0.1, 0.1\} & \{1.5, 0.55\} &  &  & \CheckmarkBold \\
5506-13-3-02-21-34-38 & \{0.1, 0.1\} & \{1.5, 0.55\} &  &  & \CheckmarkBold \\
5506-3-15-17-12-18-03 & 0.1 & 8.75 & \CheckmarkBold &  &  \\
5506-8-5-02-21-39-20 & \{0.1, 0.1\} & \{1.5, 0.55\} &  &  & \CheckmarkBold \\
5506-10-0-01-15-43-53 & 0.35 & 3.5 &  & \CheckmarkBold &  \\
5506-13-4-02-21-40-07 & 0.1 & 8.75 & \CheckmarkBold &  &  \\
5506-4-12-01-15-10-52 & 0.35 & 3.5 &  & \CheckmarkBold &  \\
5506-8-6-01-15-22-56 & \{0.1, 0.1\} & \{1.5, 0.55\} &  &  & \CheckmarkBold \\
5506-10-12-02-12-35-19 & 0.1 & 8.75 & \CheckmarkBold &  &  \\
5506-13-5-02-21-35-41 & 0.1 & 8.75 & \CheckmarkBold &  &  \\
5506-4-14-01-14-32-47 & 0.1 & 8.75 & \CheckmarkBold &  &  \\
5506-9-12-01-10-32-52 & 0.1 & 8.75 & \CheckmarkBold &  &  \\
5506-10-13-01-15-03-41 & 0.35 & 3.5 &  & \CheckmarkBold &  \\
5506-14-11-01-13-58-37 & \{0.1, 0.1\} & \{1.5, 0.55\} &  &  & \CheckmarkBold \\
5506-4-16-17-05-46-47 &  &  &  \\
5506-9-2-01-14-19-00 & \{0.1, 0.1\} & \{1.5, 0.55\} &  &  & \CheckmarkBold \\
5506-10-14-17-10-38-34 & 0.35 & 3.5 &  & \CheckmarkBold &  \\
5506-14-12-01-12-02-42 & \{0.1, 0.1\} & \{1.5, 0.55\} &  &  & \CheckmarkBold \\
5506-4-3-01-09-35-18 & 0.35 & 3.5 &  & \CheckmarkBold &  \\
5506-9-3-01-14-23-39 & \{0.1, 0.1\} & \{1.5, 0.55\} &  &  & \CheckmarkBold \\
5506-10-3-01-15-22-34 & 0.35 & 3.5 &  & \CheckmarkBold &  \\
5506-14-15-01-15-20-33 & 0.35 & 3.5 &  & \CheckmarkBold &  \\
5506-5-12-01-15-05-55 &  &  &  \\
5506-9-7-01-13-40-02 & 0.1 & 8.75 & \CheckmarkBold &  &  \\
5506-10-3-17-12-09-26 &  &  &  \\
5506-14-2-02-15-14-46 & 0.1 & 8.75 & \CheckmarkBold &  &  \\
5506-5-16-02-21-15-42 &  &  &  \\
5506-9-9-01-13-15-48 & 0.35 & 3.5 &  & \CheckmarkBold &  \\
5506-11-2-01-14-11-40 & 0.35 & 3.5 &  & \CheckmarkBold &  \\
5506-14-5-01-15-59-52 &  &  &  \\
5506-5-3-02-18-52-53 & \{0.1, 0.1\} & \{1.5, 0.55\} &  &  & \CheckmarkBold \\
5506-11-4-17-12-33-10 & \{0.1, 0.1\} & \{1.5, 0.55\} &  &  & \CheckmarkBold \\
5506-15-11-01-10-04-14 & 0.35 & 3.5 &  & \CheckmarkBold &  \\
5506-6-11-01-14-16-09 & 0.35 & 3.5 &  & \CheckmarkBold &  \\
5506-11-6-17-12-43-05 &  &  &  \\
5506-15-11-01-12-54-35 & \{0.1, 0.1\} & \{1.5, 0.55\} &  &  & \CheckmarkBold \\
5506-6-2-01-09-16-44 & 0.35 & 3.5 &  & \CheckmarkBold &  \\
5506-12-11-17-05-56-16 & 0.35 & 3.5 &  & \CheckmarkBold &  \\
5506-15-11-17-12-12-28 &  &  &  \\
5506-6-3-01-15-20-20 & 0.35 & 3.5 &  & \CheckmarkBold &  \\
\bottomrule
\end{tabular}
}
\end{small}
\end{center}
\caption{
Unimal IDs we adapted from \citet{gupta2022metamorph}. We inspect 100 morphologies and select 72 morphologies that work healthily. In the experiment of \autoref{tab:distillation} and \autoref{tab:unimal}, we select 20 morphologies each for 3 tasks (reach, touch, twisters) as listed above.
}
\label{tab:unimal_id}
\end{table*}

\begin{table*}[ht]
\begin{center}
\begin{small}
\scalebox{0.85}{
\begin{tabular}{llcccc}
\toprule
\textbf{Sub-domain} & \textbf{Environment} & $d_{\min}^{\psi}$ & $d_{\max}^{\psi}$ & \textbf{Task-Test} & \textbf{OOD-Test} \\
\midrule
\multirow{4}{*}{ant\_push} & ant\_push\_3 &  1.0 & 4.0 & \CheckmarkBold &  \\ 
& ant\_push\_4 &  1.0 & 4.0 & \CheckmarkBold &  \\ 
& ant\_push\_5 &  1.0 & 4.0 & \CheckmarkBold &  \\ 
& ant\_push\_6 &  1.0 & 4.0 & \CheckmarkBold &  \\ 
\midrule
\multirow{6}{*}{centipede\_push} & centipede\_push\_3 & 1.5 & 3.75 & \CheckmarkBold &  \\ 
& centipede\_push\_4 & 1.5 & 3.75 & \CheckmarkBold &  \\ 
& centipede\_push\_5 & 1.5 & 3.75 & \CheckmarkBold &  \\ 
& centipede\_push\_6 & 1.5 & 3.75 & \CheckmarkBold &  \\ 
& centipede\_push\_7 & 1.5 & 3.75 & \CheckmarkBold &  \\ 
\midrule
\multirow{3}{*}{worm\_push} & ant\_push\_3 & 1.5 & 3.25 & \CheckmarkBold &  \\ 
& worm\_push\_4 & 1.5 & 3.25 & \CheckmarkBold &  \\ 
& worm\_push\_5 & 1.5 & 3.25 & \CheckmarkBold &  \\ 
\midrule
\multirow{11}{*}{ant\_push\_diverse} & ant\_push\_4\_b &  1.0 & 4.0 & &  \CheckmarkBold \\ 
& ant\_push\_5\_b &  1.0 & 4.0 & &  \CheckmarkBold \\ 
& ant\_push\_4\_mass\_0.5\_1.0\_3.0 &  1.0 & 4.0 & &  \CheckmarkBold \\ 
& ant\_push\_4\_mass\_0.5\_1.0\_1.0 &  1.0 & 4.0 & &  \CheckmarkBold \\ 
& ant\_push\_4\_mass\_1.0\_3.0\_3.0 &  1.0 & 4.0 & &  \CheckmarkBold \\ 
& ant\_push\_4\_size\_0.9\_1.0\_1.1 &  1.0 & 4.0 & &  \CheckmarkBold \\ 
& ant\_push\_5\_mass\_0.5\_1.0\_3.0 &  1.0 & 4.0 & &  \CheckmarkBold \\ 
& ant\_push\_5\_mass\_0.5\_1.0\_1.0 &  1.0 & 4.0 & &  \CheckmarkBold \\ 
& ant\_push\_5\_mass\_1.0\_3.0\_3.0 &  1.0 & 4.0 & &  \CheckmarkBold \\ 
& ant\_push\_5\_size\_0.9\_1.0\_1.1 &  1.0 & 4.0 & &  \CheckmarkBold \\ 
\midrule
\multirow{17}{*}{centipede\_push\_diverse} & centipede\_push\_3\_b\_r\_0 & 1.5 & 3.75 & & \CheckmarkBold \\
& centipede\_push\_3\_b\_l\_0 & 1.5 & 3.75 & &  \CheckmarkBold \\ 
& centipede\_push\_3\_b\_r\_1 & 1.5 & 3.75 & &  \CheckmarkBold \\ 
& centipede\_push\_3\_b\_l\_1 & 1.5 & 3.75 & &  \CheckmarkBold \\ 
& centipede\_push\_4\_b\_r\_0 & 1.5 & 3.75 & &  \CheckmarkBold \\ 
& centipede\_push\_4\_b\_r\_1 & 1.5 & 3.75 & &  \CheckmarkBold \\ 
& centipede\_push\_4\_b\_l\_1 & 1.5 & 3.75 & &  \CheckmarkBold \\ 
& centipede\_push\_4\_b\_r\_2 & 1.5 & 3.75 & &  \CheckmarkBold \\ 
& centipede\_push\_4\_b\_l\_2 & 1.5 & 3.75 & &  \CheckmarkBold \\ 
& centipede\_push\_3\_size\_0.9\_1.0\_1.1 & 1.5 & 3.75 & &  \CheckmarkBold \\ 
& centipede\_push\_3\_mass\_0.5\_1.0\_3.0 & 1.5 & 3.75 & &  \CheckmarkBold \\ 
& centipede\_push\_3\_mass\_0.5\_1.0\_1.0 & 1.5 & 3.75 & &  \CheckmarkBold \\ 
& centipede\_push\_3\_mass\_1.0\_3.0\_3.0 & 1.5 & 3.75 & &  \CheckmarkBold \\ 
& centipede\_push\_4\_size\_0.9\_1.0\_1.1 & 1.5 & 3.75 & &  \CheckmarkBold \\ 
& centipede\_push\_4\_mass\_0.5\_1.0\_3.0 & 1.5 & 3.75 & &  \CheckmarkBold \\ 
& centipede\_push\_4\_mass\_0.5\_1.0\_1.0 & 1.5 & 3.75 & &  \CheckmarkBold \\
& centipede\_push\_4\_mass\_1.0\_3.0\_3.0 & 1.5 & 3.75 & &  \CheckmarkBold \\
\bottomrule
\end{tabular}
}
\end{small}
\end{center}
\caption{The extra combinations of environments used in the experiments of compositional generalization for task and out-of-distribution generalization with push task.}
\label{tab:zs_task_ood_entry_push}
\end{table*}


\clearpage
\section{Architecture Selection: Tokenized \proposedREP{}}
\label{sec:gato_representtion}

In the recent literature, offline RL is considered as supervised sequential modeling problem~\citep{chen2021decision}, and some works~\citep{janner2021sequence,reed2022gato} tokenize the continuous observations and actions, like an analogy of vision transformer~\citep{dosovitskiy2020vit}; treating input modality like a language.

As a part of offline architecture selection, we examine the effectiveness of tokenization.
We mainly follow the protocol in \citet{reed2022gato};
we first apply mu-law encoding~\citep{oord2016wavenet} to \proposedrep{} representation:
\begin{equation}
    \text{mu\_law}(x) := \text{sgn}(x)\frac{\log(|x|\mu + 1)}{\log(M\mu + 1)}, \nonumber
\end{equation}
with $\mu = 100$ and $M = 256$. This pre-processing could normalize the input to the range of $[0, 1]$. Then, we discretize pre-processed observations and actions with 1024 bins.
To examine the broader range of design choice, we prepare the following 6 variants of tokenized \proposedrep{}: whether layer normalization is added after embedding function (LN)~\citep{bhatt2019crossnorm,parisotto2019stabilize,furuta2021coadapt,chen2021decision}, or outputting discretized action (D), smoothed action by taking the average of bins (DA), and continuous values directly (C).

As shown in \autoref{tab:gato}, predicting continuous value reveals the better performance than discretized action maybe due to some approximation errors.
However, the performance of Token-\proposedshort{} (C) or Token-\proposedshort{} (C, LN) is still lower than \proposedshort{} itself.
This might be because tokenization looses some morphological invariance among nodes.
In \autoref{tab:distillation}, we adopt Token-\proposedshort{} (C) for comparison.


\begin{table*}[ht]
\begin{center}
\begin{small}
\scalebox{0.55}{
\begin{tabular}{l|cccccc|c}
\toprule
\textbf{Sub-domain} & \textbf{Token-\proposedshort{} (D, LN)} & \textbf{Token-\proposedshort{} (DA, LN)} & \textbf{Token-\proposedshort{} (C, LN)} & \textbf{Token-\proposedshort{} (D)} & \textbf{Token-\proposedshort{} (DA)} & \textbf{Token-\proposedshort{} (C)} &  \textbf{Transformer (\proposedshort{})} \\
\midrule
ant\_reach & 0.9256 $\pm$ 0.01 & 0.9252 $\pm$ 0.01 & 0.4394 $\pm$ 0.01 & 0.9210 $\pm$ 0.02 & 0.9215 $\pm$ 0.00 & 0.3846 $\pm$ 0.03 & 0.3206 $\pm$ 0.06 \\
ant\_touch & 1.0373 $\pm$ 0.01 & 1.0631 $\pm$ 0.01 & 0.4465 $\pm$ 0.01 & 1.0528 $\pm$ 0.00 & 1.0630 $\pm$ 0.01 & 0.3458 $\pm$ 0.03 & 0.2668 $\pm$ 0.08 \\
ant\_twisters & 0.5581 $\pm$ 0.00 & 0.5655 $\pm$ 0.00 & 0.2090 $\pm$ 0.00 & 0.5587 $\pm$ 0.01 & 0.5655 $\pm$ 0.00 & 0.2487 $\pm$ 0.01 & 0.1039 $\pm$ 0.05 \\
claw\_reach & 0.9568 $\pm$ 0.00 & 0.9584 $\pm$ 0.01 & 0.3685 $\pm$ 0.01 & 0.9674 $\pm$ 0.00 & 0.9583 $\pm$ 0.01 & 0.2862 $\pm$ 0.07 & 0.3581 $\pm$ 0.04 \\
claw\_touch & 1.0300 $\pm$ 0.01 & 1.0584 $\pm$ 0.01 & 0.3238 $\pm$ 0.07 & 1.0392 $\pm$ 0.03 &  1.0585 $\pm$ 0.01 & 0.3229 $\pm$ 0.07 & 0.2573 $\pm$ 0.08 \\
claw\_twisters & 0.6228 $\pm$ 0.01 & 0.6205 $\pm$ 0.00 & 0.4035 $\pm$ 0.02 & 0.6241 $\pm$ 0.00 & 0.6202 $\pm$ 0.00 & 0.3810 $\pm$ 0.03 & 0.3442 $\pm$ 0.04 \\
centipede\_reach & 0.5692 $\pm$ 0.12 & 0.5784 $\pm$ 0.13 & 0.1166 $\pm$ 0.00 & 0.6373 $\pm$ 0.09 & 0.5780 $\pm$ 0.13 & 0.1132 $\pm$ 0.03 & 0.1057 $\pm$ 0.04 \\
centipede\_touch &0.9818 $\pm$ 0.01 & 1.0088 $\pm$ 0.00 & 0.1696 $\pm$ 0.00 & 0.9910 $\pm$ 0.01 & 1.0087 $\pm$ 0.00 & 0.1823 $\pm$ 0.03  & 0.3869 $\pm$ 0.04 \\
worm\_touch & 1.0596 $\pm$ 0.01 & 1.0731 $\pm$ 0.01 & 0.9357 $\pm$ 0.14 & 1.0639 $\pm$ 0.01 & 1.0706 $\pm$ 0.01 & 0.9235 $\pm$ 0.07 & 0.8952 $\pm$ 0.04 \\
\midrule
\textbf{Average Dist.}  & 0.8124 $\pm$ 0.01 & 0.8232 $\pm$ 0.02 & 0.3572 $\pm$ 0.02 & 0.8248 $\pm$ 0.01 & 0.8225 $\pm$ 0.02 & 0.3402 $\pm$ 0.01 & 0.3128 $\pm$ 0.02 \\
\bottomrule
\end{tabular}
}
\end{small}
\end{center}
\caption{The average normalized final distance in in-distribution evaluation.
We extensively evaluate the tokenized \proposedrep{} variants, similar to \citet{reed2022gato}.
}
\label{tab:gato}
\end{table*}

\clearpage

\section{Architecture Selection: Position Embedding}
\label{sec:pe_results}
As a part of architecture selection, we investigate whether position embedding (PE) contributes to the generalization.
Our empirical results in \autoref{tab:pos} suggest that, multi-task goal reaching performance seems comparable between those. However, in more diverse morphology domains, PE plays an important role.
Therefore, we include PE into a default design.

\begin{table*}[ht]
\begin{center}
\begin{small}
\scalebox{0.825}{
\begin{tabular}{l|cc}
\toprule
\textbf{Sub-domain} & \textbf{Transformer (\proposedshort{}) (w/ PE)} & \textbf{Transformer (\proposedshort{}) (w/o PE)} \\
\midrule
ant\_reach & 0.3206 $\pm$ 0.06 & 0.3966 $\pm$ 0.08\\
ant\_touch & 0.2668 $\pm$ 0.08 & 0.4573 $\pm$ 0.14 \\
ant\_twisters & 0.1039 $\pm$ 0.05 & 0.1569 $\pm$ 0.02 \\
claw\_reach & 0.3581 $\pm$ 0.04 & 0.3508 $\pm$ 0.03 \\
claw\_touch & 0.2573 $\pm$ 0.08 & 0.3278 $\pm$ 0.07 \\
claw\_twisters & 0.3442 $\pm$ 0.04 & 0.3071 $\pm$ 0.04\\
centipede\_reach & 0.1057 $\pm$ 0.04 & 0.0610 $\pm$ 0.02 \\
centipede\_touch & 0.3869 $\pm$ 0.04 & 0.1687 $\pm$ 0.03 \\
worm\_touch & 0.8952 $\pm$ 0.05 & 0.8427 $\pm$ 0.03\\
\midrule
\textbf{Average Dist.} & 0.3128 $\pm$ 0.02  & 0.3142 
$\pm$ 0.03 \\
\midrule
unimal\_reach & 0.4532 $\pm$ 0.01 & 0.5856 $\pm$ 0.02 \\
unimal\_touch & 0.4461 $\pm$ 0.05 & 0.3799 $\pm$ 0.06 \\
unimal\_twisters & 0.3540 $\pm$ 0.02 & 0.3236 $\pm$ 0.04 \\
\midrule
\textbf{Average Dist.}  & 0.4178 $\pm$ 0.01 & 0.4297 $\pm$ 0.03 \\
\bottomrule
\end{tabular}
}
\end{small}
\end{center}
\caption{The average normalized final distance on in-distribution evaluation.
We compare the effect of position embedding.
}
\label{tab:pos}
\end{table*}

\updates{

\section{Representation Selection: \proposedREP{} with History}
\label{sec:temporal_history}

Some meta RL or multi-task RL algorithms take historical observations as inputs, or leverage recurrent neural networks to encode the temporal information of the tasks~\citep{wang2016metarl,teh2017distral,rakelly2019efficient}. To examine whether \proposedrep{} could be augmented with historical observations, we test the \proposedrep{} with history, which concatenates recent $H$ frames \proposedrep{} as inputs while predicting one-step actions, in various types of morphology-task generalization with ant agents.
In all the setting, we choose ant\_\{reach, touch, twisters\} (in \autoref{tab:zs_morph_entry}) for training.
In morphology generalization setting, we hold out ant\_5 agent tasks for evaluation.
In task generalization setting, we use ant\_reach\_hard (in \autoref{tab:zs_task_ood_entry}) as a test set.
In out-of-distribution setting, we use ant\_reach\_hard\_diverse (in \autoref{tab:zs_task_ood_entry}) for evaluation. We also set the history length as $H=3$.

\autoref{tab:history_ant} implies that when the policy faces unseen tasks (i.e. in Compositional (Task) or Out-of-Distribution settings), \proposedrep{} with history may help to improve the performance, which seems promising results to extend our framework to more complex tasks.

}

\begin{table*}[ht]
\begin{center}
\begin{small}
\scalebox{0.8}{
\begin{tabular}{l|cccccc}
\toprule
 & \textbf{MLP} & \textbf{Transformer (\proposedvariant{})} & \textbf{Transformer (\proposedshort{})} & \textbf{Transformer (\proposedshort{}-history)} \\
\midrule
\textbf{In-Distribution} & 0.3035 $\pm$ 0.03 & 0.1364 $\pm$ 0.04 & 0.0961 $\pm$ 0.02 & 0.1015 $\pm$ 0.04 \\
\midrule
\textbf{Compositional (Morphology)} & 0.7480 $\pm$ 0.02 & 0.1477 $\pm$ 0.04 & 0.1010 $\pm$ 0.01 & 0.1070 $\pm$ 0.03 \\
\textbf{Compositional (Task)} & 0.4468 $\pm$ 0.11 & 0.2243 $\pm$ 0.03 & 0.2201 $\pm$ 0.04 & 0.2170 $\pm$ 0.01 \\
\textbf{Out-of-Distribution} & 0.6788 $\pm$ 0.01 & 0.3663 $\pm$ 0.05 & 0.3370 $\pm$ 0.03 & 0.3316 $\pm$ 0.05 \\
\bottomrule
\end{tabular}
}
\end{small}
\end{center}
\caption{\updates{The average normalized final distance in various types of morphology-task generalization on \proposedbench{} (especially, ant).
We concatenate recent three frames \proposedrep{} as inputs to Transformer.
The results show that when the policy faces unseen tasks, \proposedshort{}-history may help to improve the performance.
}}
\label{tab:history_ant}
\end{table*}

\clearpage


\updates{
\section{Does Transformer with \proposedREP{} scale up with dataset/morphology-task/model size?}
\label{sec:data_entry}
}

The important aspect of the success in large language models is the scalability to the size of training data, the number of tasks for joint-training, \updates{and the number of parameters}~\citep{2020t5,brown2020language}.
One natural question is whether a similar trend holds even in RL.

\begin{figure*}[ht]
\centering
\includegraphics[width=1\linewidth]{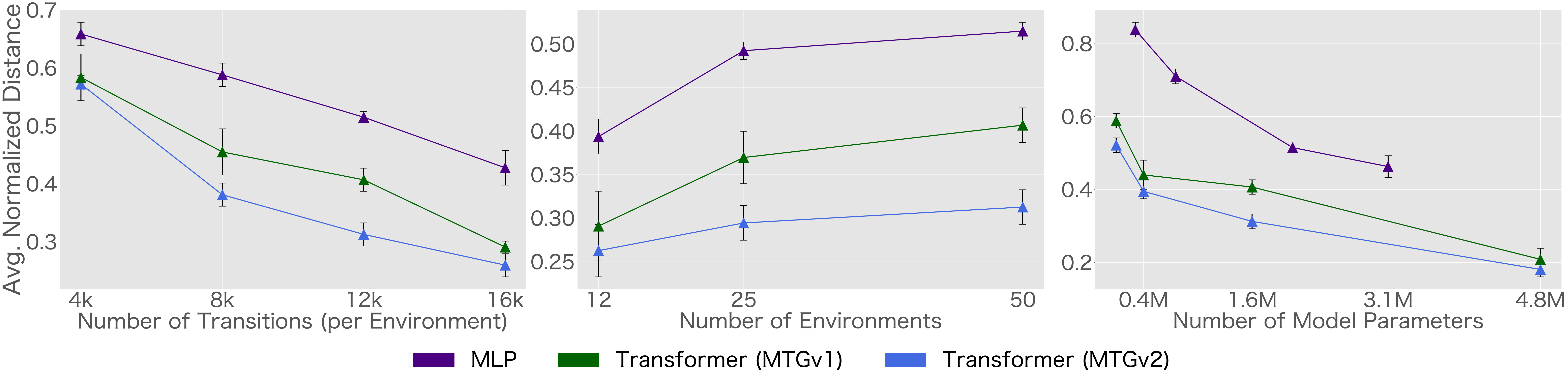}
\vspace{-5pt}
\caption{
The average normalized final distance with different size of datasets (left), morphology-task combinations (middle), \updates{and model size (right)}.
The smaller value means the better multi-task performance (see \autoref{sec:environment_division} for environment division).
These results suggest that the performance can get better when we increase the number of datasets, and \proposedrep{} can surpasses the degradation of the performance when we increase the number of environments.
\updates{Transformer is parameter-efficient than MLP, and improves the performance as many parameters.}
}
\label{fig:data_entry}
\end{figure*}

\autoref{fig:data_entry} suggests that the performance can get better if we increase the number of datasets \updates{and the number of parameters.
The performance of Transformer with 0.4M parameters is equivalent to that of MLP with 3.1M.}
In contrast, when we increase the number of environments, the performance degrades while Transformer with \proposedvariant{} and \proposedshort{} surpasses the degree of degradation, which seems inevitable trends in multi-task RL~\citep{yu2020gradient,kurin2022indefence} and an important future direction towards generalist controllers.

\section{Percentage of Improvement}
For clarification, we compute the percentage of improvement between two average normalized final distances (defined in \autoref{eq:normalized_dist}) $\bar{d_1}$ and $\bar{d_2}$ as follows ($\bar{d_1} < \bar{d_2}$):
\begin{equation}
    100 * \frac{\bar{d_2} - \bar{d_1}}{\bar{d_2}}. \nonumber
\end{equation}

\clearpage
\section{Additional Results}
\label{sec:additional_results}
In this section, we provide the detailed performance of in-distribution generalization~(\autoref{tab:in_distillation} and \autoref{tab:unimal}), compositional morphology and task generalization~(\autoref{tab:zeroshot_morph} and \autoref{tab:zeroshot_task}), and out-of-distribution generalization~(\autoref{tab:zeroshot_ood}).
For fine-tuning experiments, we summarized the detailed scores of \autoref{fig:finetuning_all} in \autoref{tab:finetuning_details}.

\begin{table*}[ht]
\begin{center}
\begin{small}
\scalebox{0.825}{
\begin{tabular}{l|ccccc}
\toprule
\textbf{Sub-domain} & \textbf{Random} & \textbf{MLP} & \textbf{GNN (\proposedvariant{})} & \textbf{Transformer (\proposedvariant{})} & \textbf{Transformer (\proposedshort{})} \\
\midrule
ant\_reach & 0.9637 $\pm$ 0.02 &  0.6426 $\pm$ 0.03 & 0.6240 $\pm$ 0.03 & 0.3657 $\pm$ 0.04 & 0.3206 $\pm$ 0.06 \\
ant\_touch & 1.0817 $\pm$ 0.02 &  0.3689 $\pm$ 0.02 & 0.4434 $\pm$ 0.03 & 0.1140 $\pm$ 0.03 &  0.2668 $\pm$ 0.08 \\
ant\_twisters & 0.9113 $\pm$ 0.84 &  0.3708 $\pm$ 0.01 & 0.2513 $\pm$ 0.01  & 0.2517 $\pm$ 0.02 & 0.1039 $\pm$ 0.05 \\
claw\_reach & 1.0760 $\pm$ 0.20 &  0.6617 $\pm$ 0.00 & 0.6214 $\pm$ 0.02 & 0.7158 $\pm$ 0.03 & 0.3581 $\pm$ 0.04 \\
claw\_touch & 1.5240 $\pm$ 0.88 &  0.6824 $\pm$ 0.06 & 0.3135 $\pm$ 0.06 & 0.6121 $\pm$ 0.04 &  0.2573 $\pm$ 0.08 \\
claw\_twisters & 1.3786 $\pm$ 1.42 &  0.4907 $\pm$ 0.03 & 0.6063 $\pm$ 0.07  & 0.4614 $\pm$ 0.07 & 0.3442 $\pm$ 0.04 \\
centipede\_reach & 0.5843 $\pm$ 0.35 &  0.0803 $\pm$ 0.02 & 0.1088 $\pm$ 0.02 & 0.0981 $\pm$ 0.03 &  0.1057 $\pm$ 0.04 \\
centipede\_touch & 1.0077 $\pm$ 0.01 &  0.4743 $\pm$ 0.03 & 0.5089 $\pm$ 0.01 & 0.4609 $\pm$ 0.07 & 0.3869 $\pm$ 0.04 \\
worm\_touch & 2.7087 $\pm$ 1.73 &  1.1034 $\pm$ 0.06 & 1.0559 $\pm$ 0.04 & 0.7708 $\pm$ 0.03 &  0.8952 $\pm$ 0.05 \\
\midrule
\textbf{Average Dist.} & 1.2019 $\pm$ 0.41 &  0.5150 $\pm$ 0.01 & 0.4776 $\pm$ 0.01 & 0.4069 $\pm$ 0.02 & \textbf{0.3128 $\pm$ 0.02} \\
\bottomrule
\end{tabular}
}
\end{small}
\end{center}
\caption{
The average normalized final distance for in-distribution evaluation on \proposedbench{} (as shown in \autoref{tab:distillation}).
}
\label{tab:in_distillation}
\end{table*}

\begin{table*}[ht]
\begin{center}
\begin{small}
\scalebox{0.825}{
\begin{tabular}{l|cccc}
\toprule
\textbf{Sub-domain} & \textbf{Random} & \textbf{MLP} & \textbf{Transformer (\proposedvariant{})} & \textbf{Transformer (\proposedshort{})} \\
\midrule
unimal\_reach & 0.9662 $\pm$ 0.01 & 0.7448 $\pm$ 0.02 & 0.5692 $\pm$ 0.01 & 0.4532 $\pm$ 0.01 \\
unimal\_touch & 1.1302 $\pm$ 0.10 & 0.7634 $\pm$ 0.03 & 0.5290 $\pm$ 0.04 & 0.4461 $\pm$ 0.05 \\
unimal\_twisters & 0.6305 $\pm$ 0.02 & 0.5027 $\pm$ 0.02 & 0.3534 $\pm$ 0.03 & 0.3540 $\pm$ 0.02 \\
\midrule
\textbf{Average Dist.}  & 0.9090 $\pm$ 0.03 & 0.6703 $\pm$ 0.01 & 0.4839 $\pm$ 0.02 & \textbf{0.4178 $\pm$ 0.01} \\
\bottomrule
\end{tabular}
}
\end{small}
\end{center}
\caption{
The average normalized final distance for in-distribution evaluation on \proposedbench{} with challenging morphologies from \citet{gupta2022metamorph} (as shown in \autoref{tab:distillation}).
}
\label{tab:unimal}
\end{table*}

\begin{table*}[ht]
\begin{center}
\begin{small}
\scalebox{0.825}{
\begin{tabular}{l|cccc}
\toprule
\textbf{Sub-domain} & \textbf{Random} & \textbf{MLP} & \textbf{Transformer (\proposedvariant{})} & \textbf{Transformer (\proposedshort{})} \\
\midrule
ant\_reach & 0.9697 $\pm$ 0.03 & 0.8541 $\pm$ 0.03 & 0.5511 $\pm$ 0.10 & 0.4170 $\pm$ 0.04 \\
ant\_touch & 1.0821 $\pm$ 0.00 & 0.7742 $\pm$ 0.07 & 0.4464 $\pm$ 0.07 & 0.3752 $\pm$ 0.12 \\
ant\_twisters & 0.9215 $\pm$ 0.95 & 0.5356 $\pm$ 0.02 & 0.2569 $\pm$ 0.04 & 0.2608 $\pm$ 0.02 \\
claw\_reach & 0.9915 $\pm$ 0.04 & 0.9370 $\pm$ 0.01 & 0.7332 $\pm$ 0.03 & 0.4399 $\pm$ 0.06 \\
claw\_touch & 1.0695 $\pm$ 0.02 & 1.0283 $\pm$ 0.02 & 0.7074 $\pm$ 0.10 & 0.1375 $\pm$ 0.01 \\
claw\_twisters & 0.7001 $\pm$ 0.17 & 0.5456 $\pm$ 0.02 & 0.5486 $\pm$ 0.02 & 0.5589 $\pm$ 0.05 \\
centipede\_reach & 0.7929 $\pm$ 0.04 & 0.4626 $\pm$ 0.15 & 0.2640 $\pm$ 0.07 & 0.2506 $\pm$ 0.14 \\
centipede\_touch & 1.0121 $\pm$ 0.01 & 0.7632 $\pm$ 0.02 & 0.4971 $\pm$ 0.06 & 0.3474 $\pm$ 0.09 \\
worm\_touch & 2.7376 $\pm$ 2.23 & 1.1417 $\pm$ 0.10 & 0.8608 $\pm$ 0.07 & 1.0110 $\pm$ 0.03 \\
\midrule
\textbf{Average Dist.}  & 1.1419 $\pm$ 0.41 & 0.7216 $\pm$ 0.01 & 0.4940 $\pm$ 0.01 & \textbf{0.4066 $\pm$ 0.01} \\
\bottomrule
\end{tabular}
}
\end{small}
\end{center}
\caption{
The average normalized final distance for compositional morphology evaluation on \proposedbench{} (as shown in \autoref{tab:distillation}).
}
\label{tab:zeroshot_morph}
\end{table*}

\begin{table*}[ht]
\begin{center}
\begin{small}
\scalebox{0.825}{
\begin{tabular}{l|cccc}
\toprule
\textbf{Sub-domain} & \textbf{Random} & \textbf{MLP} & \textbf{Transformer (\proposedvariant{})} & \textbf{Transformer (\proposedshort{})} \\
\midrule
ant\_reach\_hard & 0.9542 $\pm$ 0.01 & 0.8299 $\pm$ 0.03 & 0.6176 $\pm$ 0.07 & 0.4522 $\pm$ 0.06 \\
centipede\_reach\_hard & 0.8443 $\pm$ 0.02 & 0.5689 $\pm$ 0.03 & 0.4770 $\pm$ 0.05 &  0.4412 $\pm$ 0.05 \\
\midrule
\textbf{Average Dist.}  & 0.8932 $\pm$ 0.01 & 0.6849 $\pm$ 0.01 & 0.5395 $\pm$ 0.04 & \textbf{0.4461 $\pm$ 0.05} \\
\bottomrule
\end{tabular}
}
\end{small}
\end{center}
\caption{
The average normalized final distance for compositional task evaluation on \proposedbench{} (as shown in \autoref{tab:distillation}).
}
\label{tab:zeroshot_task}
\end{table*}

\begin{table*}[ht]
\begin{center}
\begin{small}
\scalebox{0.825}{
\begin{tabular}{l|cccc}
\toprule
\textbf{Sub-domain} & \textbf{Random} & \textbf{MLP} & \textbf{Transformer (\proposedvariant{})} & \textbf{Transformer (\proposedshort{})} \\
\midrule
ant\_reach\_hard\_diverse & 0.9520 $\pm$ 0.01 & 0.8288 $\pm$ 0.02 & 0.6798 $\pm$ 0.04 & 0.5365 $\pm$ 0.05 \\
centipede\_reach\_hard\_diverse & 0.8660 $\pm$ 0.02 & 0.7546 $\pm$ 0.02 & 0.6130 $\pm$ 0.03 & 0.5208 $\pm$ 0.04 \\
\midrule
\textbf{Average Dist.}  & 0.8979 $\pm$ 0.01 & 0.7821 $\pm$ 0.02 & 0.6144 $\pm$ 0.04 & \textbf{0.5266 $\pm$ 0.04} \\
\bottomrule
\end{tabular}
}
\end{small}
\end{center}
\caption{
The average normalized final distance for out-of-distribution evaluation on \proposedbench{} (as shown in \autoref{tab:distillation}).
The agents are diversified with missing, mass, size randomization.
}
\label{tab:zeroshot_ood}
\end{table*}

\begin{table*}[ht]
\begin{center}
\begin{small}
\scalebox{0.9}{
\begin{tabular}{l|ccc}
\toprule
\textbf{Method (data size)} & \textbf{Compositional (Morphology)} & \textbf{Compositional (Task)} & \textbf{Out-of-Distribution} \\
\midrule
MLP (4K, randinit) & 0.8162 $\pm$ 0.02  & 0.7046 $\pm$ 0.01 & 0.6674 $\pm$ 0.03 \\
MLP (4K, fine-tuning) & 0.6715 $\pm$ 0.02 & 0.5582 $\pm$ 0.03 & 0.5694 $\pm$ 0.04 \\
Transformer (\proposedvariant{}) (4K, randinit) & 0.6021 $\pm$ 0.03 & 0.5017 $\pm$ 0.04 & 0.5222  $\pm$ 0.06 \\
Transformer (\proposedvariant{}) (4K, fine-tuning) & 0.3499 $\pm$ 0.04 & 0.3378 $\pm$ 0.08 & 0.3401 $\pm$ 0.08 \\
Transformer (\proposedshort{}) (4K, randinit) & 0.5504 $\pm$ 0.03 & 0.4894 $\pm$ 0.03 & 0.4980  $\pm$ 0.05 \\
Transformer (\proposedshort{}) (4K, fine-tuning) & 0.2863 $\pm$ 0.02 & 0.3180 $\pm$ 0.07 &  0.3115 $\pm$ 0.07 \\
\midrule
MLP (8K, randinit) & 0.7044 $\pm$ 0.02 & 0.4818 $\pm$ 0.02 & 0.5247 $\pm$ 0.04 \\
MLP (8K, fine-tuning) & 0.5579 $\pm$ 0.06 & 0.3709 $\pm$ 0.01 & 0.4238 $\pm$ 0.05 \\
Transformer (\proposedvariant{}) (8K, randinit) & 0.4001 $\pm$ 0.04 & 0.2791 $\pm$ 0.06 & 0.2762  $\pm$ 0.04 \\
Transformer (\proposedvariant{}) (8K, fine-tuning) & 0.2600 $\pm$ 0.04 & 0.1926 $\pm$ 0.03 & 0.2089 $\pm$ 0.03 \\
Transformer (\proposedshort{}) (8K, randinit) & 0.3758  $\pm$ 0.03 & 0.1868 $\pm$ 0.05 & 0.1868 $\pm$ 0.05 \\
Transformer (\proposedshort{}) (8K, fine-tuning) & 0.1871 $\pm$ 0.01 & 0.1356 $\pm$ 0.03 &  0.1498 $\pm$ 0.03 \\
\midrule
MLP (12K, randinit) &  0.5530 $\pm$ 0.01 & 0.4256 $\pm$ 0.03 & 0.4399  $\pm$ 0.02 \\
MLP (12K, fine-tuning) &  0.4312 $\pm$ 0.03 & 0.2982 $\pm$ 0.02 & 0.3358 $\pm$ 0.02 \\
Transformer (\proposedvariant{}) (12K, randinit) & 0.2914 $\pm$ 0.03 &  0.1555 $\pm$ 0.04 & 0.1759 $\pm$ 0.03 \\
Transformer (\proposedvariant{}) (12K, fine-tuning) & 0.2042 $\pm$ 0.01 & 0.0888 $\pm$ 0.01 & 0.0894 $\pm$ 0.01 \\
Transformer (\proposedshort{}) (12K, randinit) & 0.2655 $\pm$ 0.03 & 0.1149 $\pm$ 0.02 & 0.1149 $\pm$ 0.02 \\
Transformer (\proposedshort{}) (12K, fine-tuning) & 0.1301 $\pm$ 0.02 & 0.0562 $\pm$ 0.02 & 0.0513 $\pm$ 0.01 \\
\bottomrule
\end{tabular}
}
\end{small}
\end{center}
\caption{
The average normalized final distance among test environments in fine-tuning settings (compositional morphology/task or out-of-distribution evaluation).
}
\label{tab:finetuning_details}
\end{table*}

\clearpage
\updates{
\section{Why \proposedREP{} Works Well?: Attention Analysis}
\label{sec:attention_details}
}

\begin{figure*}[ht]
\centering
\includegraphics[width=\textwidth]{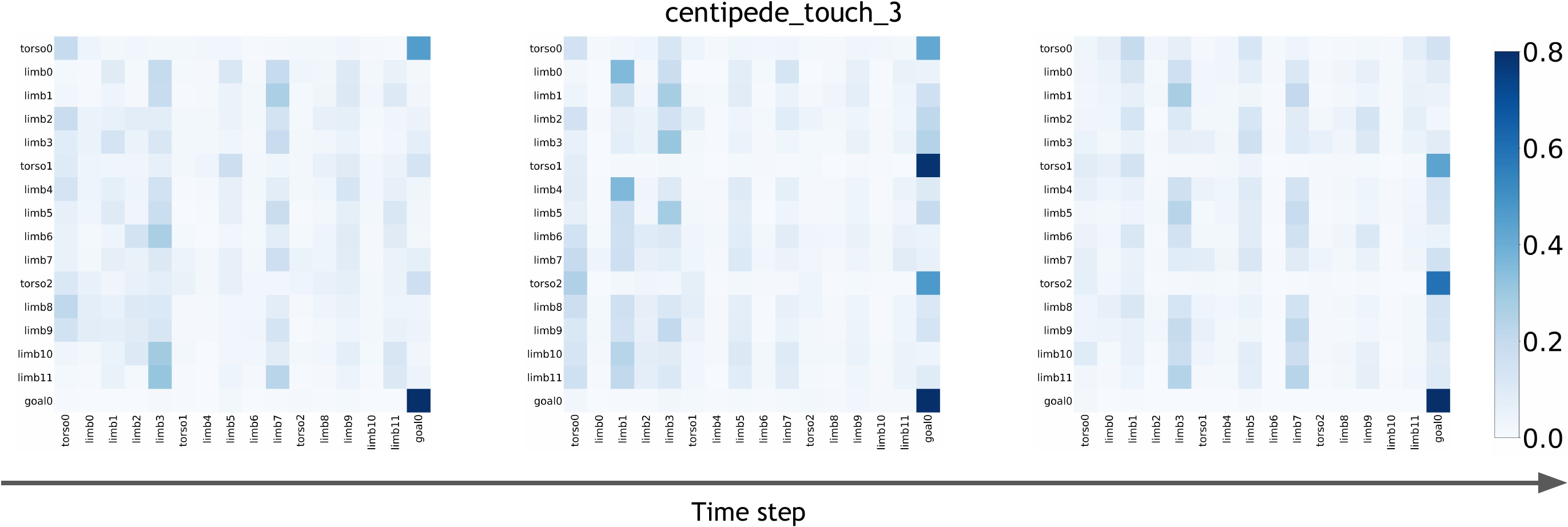}
\includegraphics[width=\textwidth]{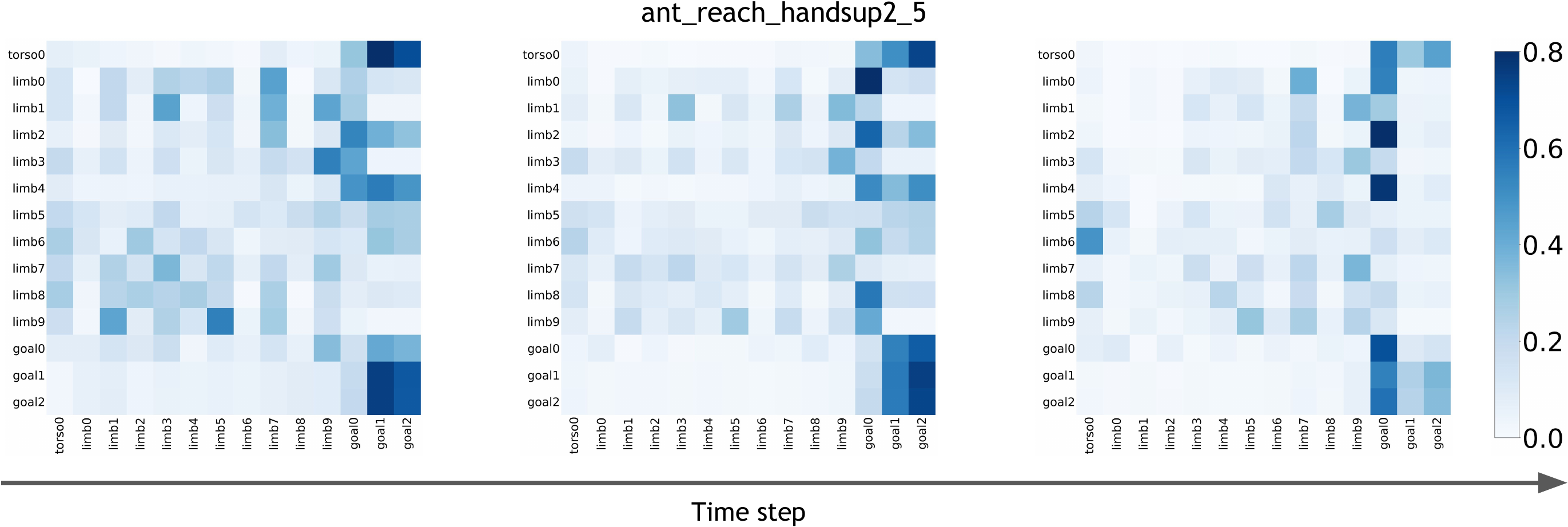}
\vspace{-5pt}
\caption{Attention analysis of \proposedshort{} in centipede\_touch\_3 (top) and ant\_reach\_handsup2\_5 (bottom; from twisters).
From left to right, we visualize the attention weights of \proposedshort{} during the rollout. In contrast to \proposedvariant{}, \proposedshort{} consistently focuses on goal nodes over time, and activates important nodes to solve the task.
}
\label{fig:atten_centipede_ant_msgt_full}
\end{figure*}

\begin{figure*}[ht]
\centering
\includegraphics[width=\textwidth]{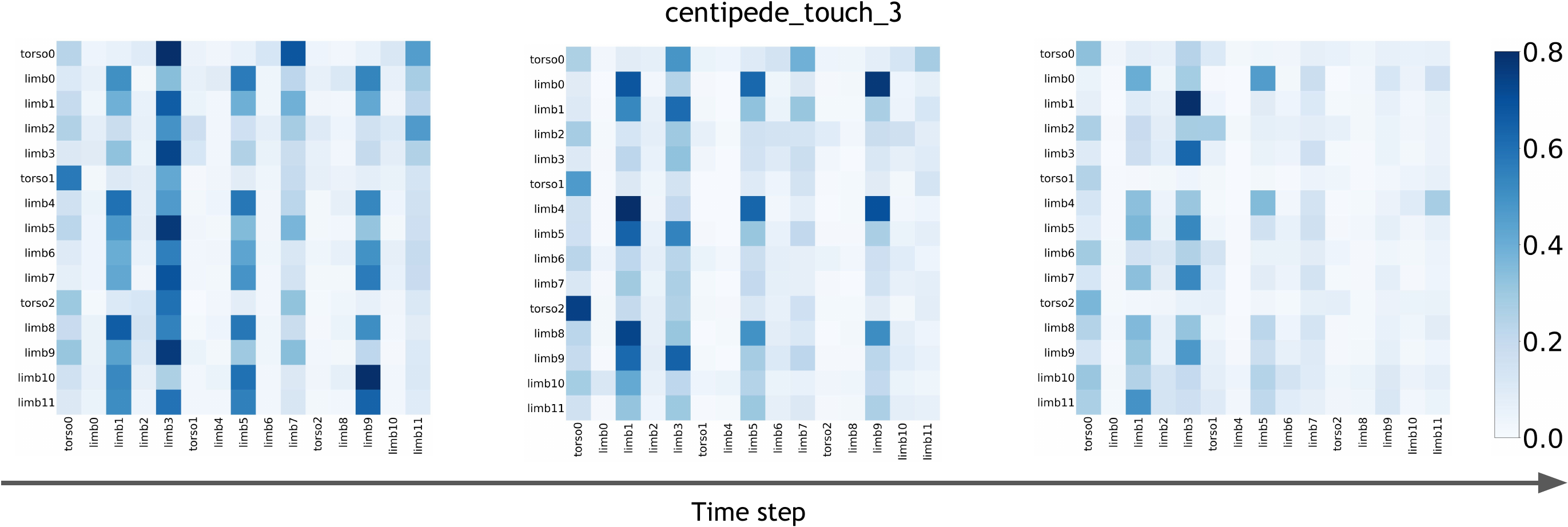}
\includegraphics[width=\textwidth]{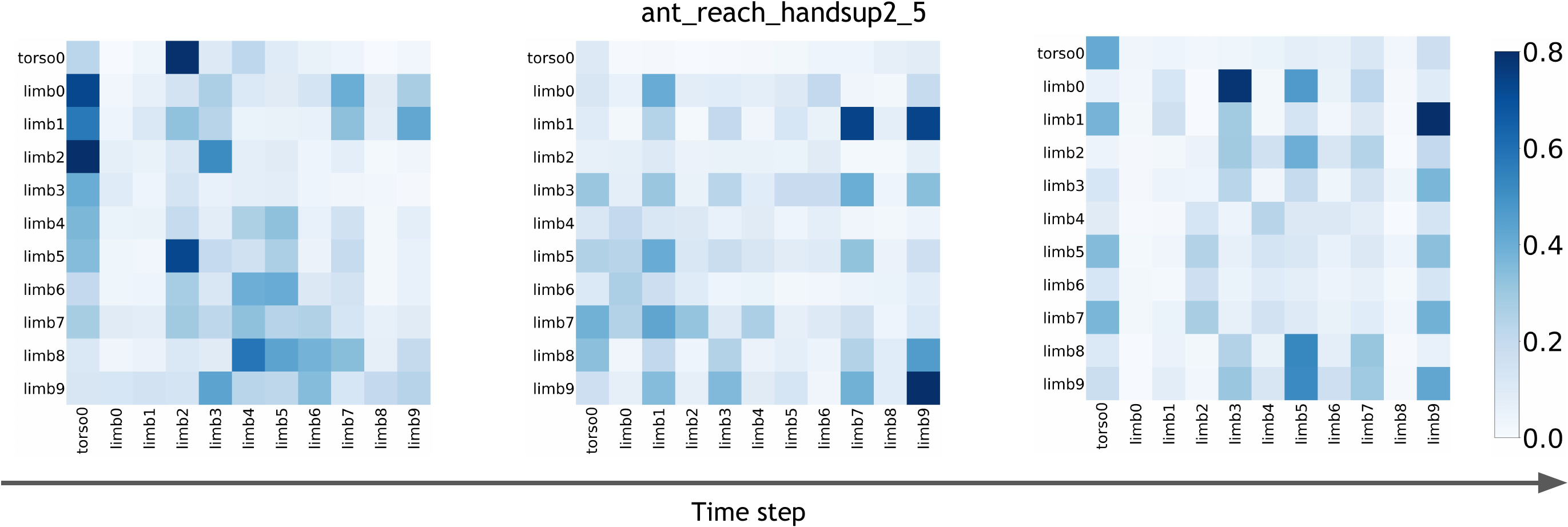}
\caption{Attention analysis of \proposedvariant{} in centipede\_touch\_3 and ant\_reach\_handsup2\_5.
This tends to demonstrate some periodic patterns during the rollout as implied in prior works~\citep{huang2020policy,kurin2020my}.
}
\label{fig:atten_centipede_ant_trans}
\end{figure*}

In this section, we provide the attention analysis of Transformer with \proposedvariant{} (\autoref{fig:atten_centipede_ant_trans}) and \proposedshort{} (\autoref{fig:atten_centipede_ant_msgt_full}).

The experimental results reveal that despite slight differences, \proposedshort{} generalizes various morphologies and tasks better than \proposedvariant{}.
To find out the difference between those, we qualitatively analyze the attention weights in Transformer.
\autoref{fig:atten_centipede_ant_msgt_full} shows that \proposedshort{} consistently focuses on goal nodes over time, and activates important nodes to solve the task; for instance, in centipede\_touch (top), \proposedshort{} pays attention to corresponding nodes (torso0 and goal0) at the beginning of the episode, and gradually sees other relevant nodes (torso1 and torso2) to hold the movable ball.
Furthermore, in ant\_twisters (bottom), \proposedshort{} firstly tries to raise the agent's legs to satisfy goal1 and goal2, and then focus on reaching a leg (goal0).
Temporally-consistent attention to goal nodes and dynamics attention to relevant nodes can contribute to generalization over goal-directed tasks and morphologies. 

\autoref{fig:atten_centipede_ant_trans} implies that \proposedvariant{} does not show such consistent activation to the goal-conditioned node; for instance, in centipede\_touch\_3 (above), the goal information is treated as an extra node feature of torso0, but there are no nodes that consistently activated with torso0.
Moreover, in ant\_reach\_handsup2\_3 (bottom), \proposedvariant{} does not keep focusing on the agent's limbs during the episode.
Rather, \proposedvariant{} tends to demonstrate some periodic patterns during the rollout as implied in prior works~\citep{huang2020policy,kurin2020my}.

\clearpage
\section{Additional Results of Task Generalization}
\label{sec:task_ood_push}

Although in \autoref{tab:distillation} and \autoref{fig:finetuning_all}, we examine the compositional task generalization and out-of-distribution generalization with reach\_hard tasks, where the goal distribution is farther than original reach tasks. While they seem more difficult unseen tasks~\citep{furuta2021pic}, they also seem to have some sort of task similarities between training dataset environments and those evaluation environments.

Another question might be how \proposedrep{} performs in more different unseen tasks.
To evaluate compositional task generalization and out-of-distribution on the environments that have less similarity to the training datasets, we prepare push task, where the agents try to move the box objects to the given goal position.
See \autoref{tab:zs_task_ood_entry_push} for the environment division. For training datasets, we leverage In-Distribution division in \autoref{tab:zs_morph_entry}.
Because this task requires the sufficient interaction with the object, the nature of tasks seem quite different from training dataset environments (reach, touch, and twisters). 

\autoref{tab:zeroshot_task_push} shows the results of compositional task evaluation and \autoref{tab:zeroshot_ood_push} shows those of out-of-distribution evaluation. In contrast to \autoref{tab:distillation}, the zero-shot performance seems limited.
Transferring pre-trained control primitives to significantly different tasks still remains as important future work.

\begin{figure}[ht]
\centering
\includegraphics[width=1.0\linewidth]{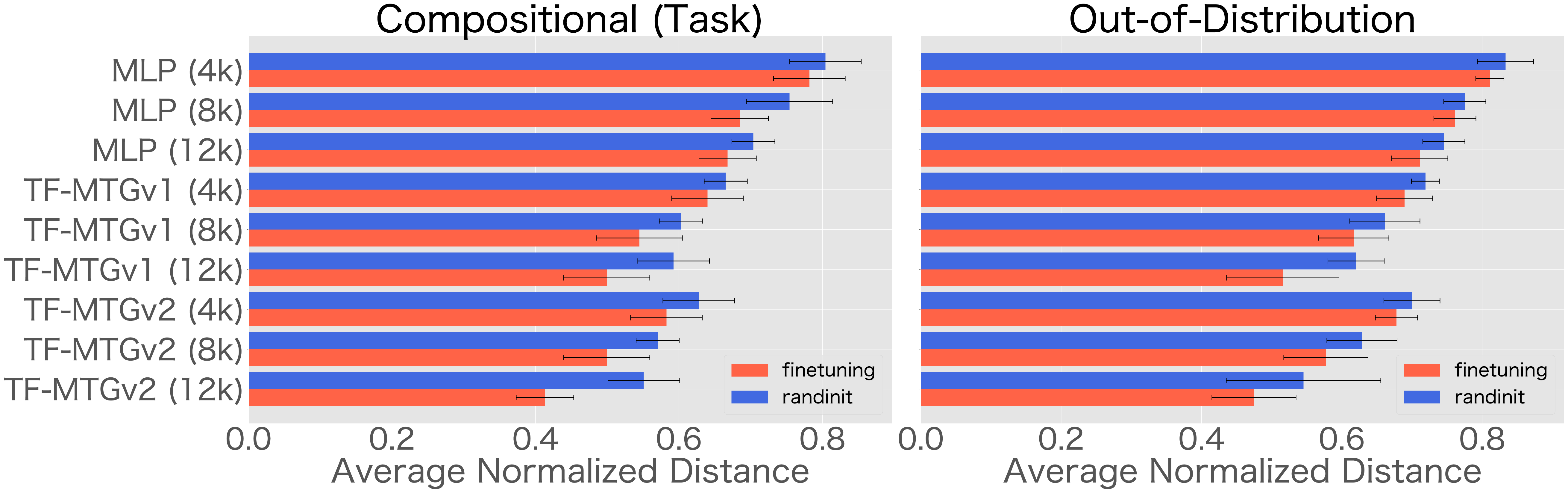}
\caption{Comparison of multi-task goal-reaching performances on fine-tuning settings in the unseen push tasks.
The results imply that \proposedrep{} works as better prior knowledge for downstream multi-task imitation learning, even with the environments that have less similarity to the pre-training datasets.
}
\label{fig:finetuning_push}
\end{figure}

However, as shown in \autoref{fig:finetuning_push}, \proposedrep{} works as better prior knowledge for downstream multi-task imitation learning, even with the environments that have less similarity to the pre-training datasets.
As prior work suggested~\citep{mandi2022finemeta}, these results suggests that, in RL, jointly-learned multi-task model has a strong inductive bias even for unseen and significantly different environments.

\begin{table*}[ht]
\begin{center}
\begin{small}
\scalebox{0.825}{
\begin{tabular}{l|cccc}
\toprule
\textbf{Sub-domain} & \textbf{Random} & \textbf{MLP} & \textbf{Transformer (\proposedvariant{})} & \textbf{Transformer (\proposedshort{})} \\
\midrule
ant\_push & 1.0808 $\pm$ 0.02 & 0.9995 $\pm$ 0.00 & 0.9212 $\pm$ 0.14  & 0.8836 $\pm$ 0.09 \\
centipede\_push & 1.0816 $\pm$ 0.01 & 0.9942 $\pm$ 0.00 & 0.9557 $\pm$ 0.02 & 0.9377 $\pm$ 0.01  \\
worm\_push & 1.1026 $\pm$ 0.04 &  0.6300 $\pm$ 0.01 & 0.5579 $\pm$ 0.10 &  0.4091 $\pm$ 0.15 \\
\midrule
\textbf{Average Dist.}  & 1.0866 $\pm$ 0.02 & 0.9051 $\pm$ 0.00 & 0.8800 $\pm$ 0.02 & 0.8263 $\pm$ 0.04 \\
\bottomrule
\end{tabular}
}
\end{small}
\end{center}
\caption{
The average normalized final distance for compositional task evaluation on \proposedbench{} with unseen push task.
See \autoref{tab:zs_task_ood_entry_push} for the environment division.
}
\label{tab:zeroshot_task_push}
\end{table*}

\begin{table*}[ht]
\begin{center}
\begin{small}
\scalebox{0.825}{
\begin{tabular}{l|cccc}
\toprule
\textbf{Sub-domain} & \textbf{Random} & \textbf{MLP} & \textbf{Transformer (\proposedvariant{})} & \textbf{Transformer (\proposedshort{})} \\
\midrule
ant\_push\_diverse & 1.0814 $\pm$ 0.02 & 1.0018 $\pm$ 0.01 & 0.9906 $\pm$ 0.01 & 0.8800 $\pm$ 0.04 \\
centipede\_push\_diverse & 1.0848 $\pm$ 0.02 & 1.0031 $\pm$ 0.01 & 0.9997 $\pm$ 0.00 & 0.9355 $\pm$ 0.04 \\
\midrule
\textbf{Average Dist.}  & 1.0835 $\pm$ 0.02 & 1.0021 $\pm$ 0.02 & 0.9965 $\pm$ 0.01 & \textbf{0.9155 $\pm$ 0.03} \\
\bottomrule
\end{tabular}
}
\end{small}
\end{center}
\caption{
The average normalized final distance for out-of-distribution evaluation on \proposedbench{} with unseen push task.
See \autoref{tab:zs_task_ood_entry_push} for the environment division.
}
\label{tab:zeroshot_ood_push}
\end{table*}

\clearpage

\begin{table*}[ht]
\begin{center}
\begin{small}
\scalebox{0.9}{
\begin{tabular}{l|cc}
\toprule
\textbf{Method (data size)} & \textbf{Compositional (Task)} & \textbf{Out-of-Distribution} \\
\midrule
MLP (4K, randinit) &  0.8042 $\pm$ 0.05 & 0.8332 $\pm$ 0.04 \\
MLP (4K, fine-tuning) & 0.7818 $\pm$ 0.05 &  0.8108 $\pm$ 0.02 \\
Transformer (\proposedvariant{}) (4K, randinit) & 0.6653 $\pm$ 0.03 & 0.7191 $\pm$ 0.02 \\
Transformer (\proposedvariant{}) (4K, fine-tuning) & 0.6398 $\pm$ 0.05 &  0.6893 $\pm$ 0.04 \\
Transformer (\proposedshort{}) (4K, randinit) & 0.6276 $\pm$ 0.05 & 0.6999 $\pm$ 0.04 \\
Transformer (\proposedshort{}) (4K, fine-tuning) & 0.5826 $\pm$ 0.05 & 0.6779 $\pm$ 0.03 \\
\midrule
MLP (8K, randinit) & 0.7542 $\pm$ 0.06 & 0.7751 $\pm$ 0.03 \\
MLP (8K, fine-tuning) & 0.6847 $\pm$ 0.04 & 0.7611 $\pm$ 0.03 \\
Transformer (\proposedvariant{}) (8K, randinit) & 0.6027 $\pm$ 0.03 & 0.6613 $\pm$ 0.05 \\
Transformer (\proposedvariant{}) (8K, fine-tuning) & 0.5448 $\pm$ 0.06 & 0.6169 $\pm$ 0.05 \\
Transformer (\proposedshort{}) (8K, randinit) & 0.5704 $\pm$ 0.03 & 0.6286 $\pm$ 0.05 \\
Transformer (\proposedshort{}) (8K, fine-tuning) & 0.4993 $\pm$ 0.06 &  0.5772 $\pm$ 0.06 \\
\midrule
MLP (12K, randinit) & 0.7037 $\pm$ 0.03 & 0.7452 $\pm$ 0.03 \\
MLP (12K, fine-tuning) & 0.6678 $\pm$ 0.04 & 0.7110 $\pm$ 0.06 \\
Transformer (\proposedvariant{}) (12K, randinit) & 0.5925 $\pm$ 0.04 & 0.6203 $\pm$ 0.04 \\
Transformer (\proposedvariant{}) (12K, fine-tuning) & 0.4993 $\pm$ 0.06 & 0.5158 $\pm$ 0.08 \\
Transformer (\proposedshort{}) (12K, randinit) & 0.5510 $\pm$ 0.05 & 0.5454 $\pm$ 0.11 \\
Transformer (\proposedshort{}) (12K, fine-tuning) & 0.4132 $\pm$ 0.04 & 0.4748 $\pm$ 0.06 \\
\bottomrule
\end{tabular}
}
\end{small}
\end{center}
\caption{
The average normalized final distance among test environments in fine-tuning settings (compositional morphology/task or out-of-distribution evaluation) with unseen push task.
}
\label{tab:finetuning_push_details}
\end{table*}


\updates{
\section{Extended Related Work}
\label{sec:extended_related_work}
}

\updates{
\textbf{Morphology-Task Generalization}~
In the previous literature, \textit{multi-task} in RL has arisen from each MDP component; from different (1) dynamics~\citep{hallak2015contextual,yu2019meta}, (2) reward~\citep{Kaelbling93learning,andrychowicz2017hindsight}, and (3) state/action space~\citep{wang2018nervenet,huang2020policy}.
We think multi-morphology RL covers (1) and (3), and multi-task (or multi-goal) RL does (2) and (1).
Considering morphology-task generalization, we could study the generalization in RL across all the MDP components (dynamics, reward, state space, action space).
While, from a broader perspective, the notion of multi-task in RL may contain both morphological and goal diversity, in this paper, we treat them separately, and \textit{multi-task} just stands for \textit{multi-goal} settings.

\textbf{Connection to Policy Distillation}~
\textit{Policy distillation} has been proposed and studied for a while~\citep{rusu2016policy,parisotto2016actor,levine2016end,Czarnecki2019distilling}, where a single student policy distills the knowledge from multiple teacher policies to obtain better multitask learners.
In contrast, we call our protocol \textit{behavior distillation}, where the single student policy distills the knowledge from \textbf{multi-source offline data} (e.g. human teleoperation, scripted behavior~\citep{singh2021parrot}, play data~\citep{lynch2019learning}), not limited to RL policy, since recent RL or robotics research often leverage such a data-driven approach for scalability~\citep{levine2020offline,chen2021a,gu2021braxlines,lee2022mgdt,reed2022gato,zeng2020transporter,shridhar2022perceiveractor}.
}

\end{document}